%% file: k-exaone.tex
\documentclass{article}

\usepackage{makecell}
\usepackage[final]{lg}
\usepackage[utf8]{inputenc} 
\usepackage[T1]{fontenc}    
\usepackage[pdfencoding=auto, psdextra, unicode]{hyperref}       
\usepackage{url}            
\usepackage{booktabs}       
\usepackage{rotating}
\usepackage{amsfonts}       
\usepackage{amsmath}
\usepackage{nicefrac}       
\usepackage[dvipsnames]{xcolor}
\usepackage{kotex}
\usepackage{adjustbox}
\usepackage{multirow}
\usepackage{amssymb}
\usepackage{longtable}
\usepackage{comment}
\usepackage[english]{babel}
\usepackage[autostyle]{csquotes}
\usepackage{hhline}
\usepackage{tabu}
\usepackage{tablefootnote}
\usepackage{longtable}
\usepackage{graphicx}
\usepackage{wrapfig}
\usepackage{array}
\usepackage{caption}
\usepackage{tcolorbox}
\usepackage{lipsum}
\usepackage{relsize}
\usepackage{colortbl}
\usepackage{tablefootnote}
\tcbuselibrary{breakable}
\usepackage{floatpag}
\floatpagestyle{plain}
\usepackage{float}
\usepackage{threeparttable}
\usepackage{enumitem}
\setlist[enumerate,itemize]{leftmargin=1.5em}
\usepackage{tcolorbox}
\usepackage{placeins}
\usepackage{setspace}
\usepackage{wrapstuff}

\newtcolorbox{HeroCard}{
  breakable,
  colback=black!3,
  colframe=black!3,
  boxrule=0pt,
  arc=3mm,
  left=12pt,right=12pt,top=10pt,bottom=0pt
}

\definecolor{MainPurple}{HTML}{A451E4}

\newcommand{\PaperHuggingFaceURL}{https://huggingface.co/LGAI-EXAONE/K-EXAONE-2.0-750B-A37B}
\newcommand{\PaperGitHubURL}{https://github.com/LG-AI-EXAONE/K-EXAONE-2.0}

\makeatletter
\edef\OrigRM{\rmdefault}
\edef\OrigSF{\sfdefault}
\edef\OrigTT{\ttdefault}
\renewcommand{\@maketitle}{

  \begin{center}
  \begin{HeroCard}
  {
    \fontfamily{put}\selectfont  

    {\LARGE\bfseries \@title\par}
    \vspace{0.5em}

    {\normalsize \@author\par}
    \vspace{0.5em}

    {\normalsize 
    \setlength{\baselineskip}{1.25\baselineskip} 
    \noindent \input{sections/00_abstract}\par}

    \vspace{1.0em}
    \noindent
    \begin{minipage}[b]{0.78\linewidth}
    {\small
      \noindent\textbf{GitHub:} \href{\PaperGitHubURL}{\nolinkurl{\PaperGitHubURL}}\par
      \noindent\textbf{Hugging Face:} \href{\PaperHuggingFaceURL}{\nolinkurl{\PaperHuggingFaceURL}}\par
    }
    \end{minipage}\hfill
    \begin{minipage}[b]{0.18\linewidth}
    \raggedleft
    \raisebox{-0.3\height}{\includegraphics[height=14mm]{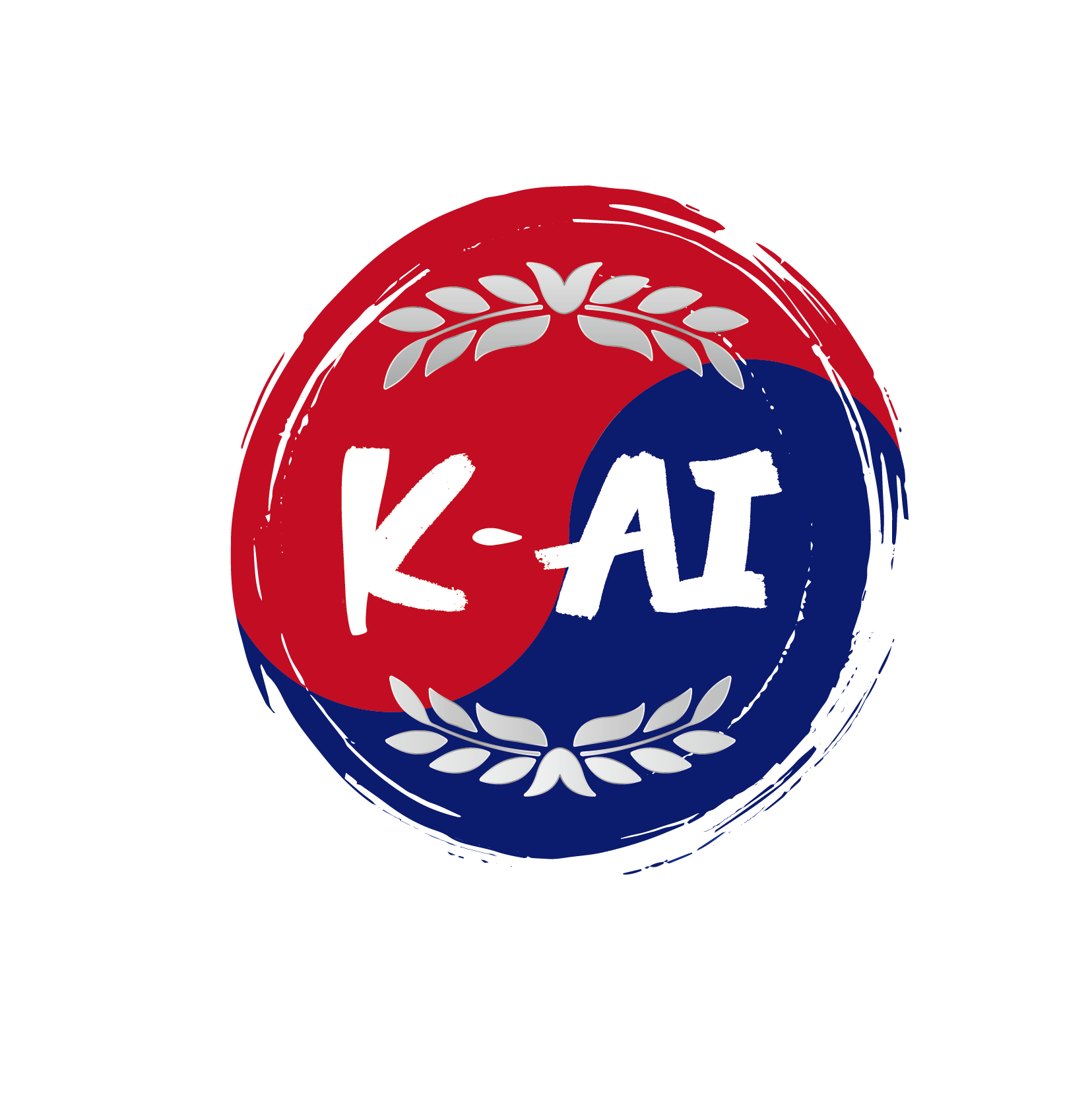}}
    \end{minipage}
  }
  \end{HeroCard}
  \end{center}

  \@thanks
  \global\let\@thanks\@empty
  \setcounter{footnote}{0}
  \vspace{0.1 em}

}
\makeatother

\usepackage{fourier} 
\renewcommand{\rmdefault}{\OrigRM}
\renewcommand{\sfdefault}{\OrigSF}
\renewcommand{\ttdefault}{\OrigTT}

\newcommand{\model}{K-EXAONE~2.0 }
\newcommand{\comp}{LG~AI~Research}

\title{
  \includegraphics[height=5.5mm]{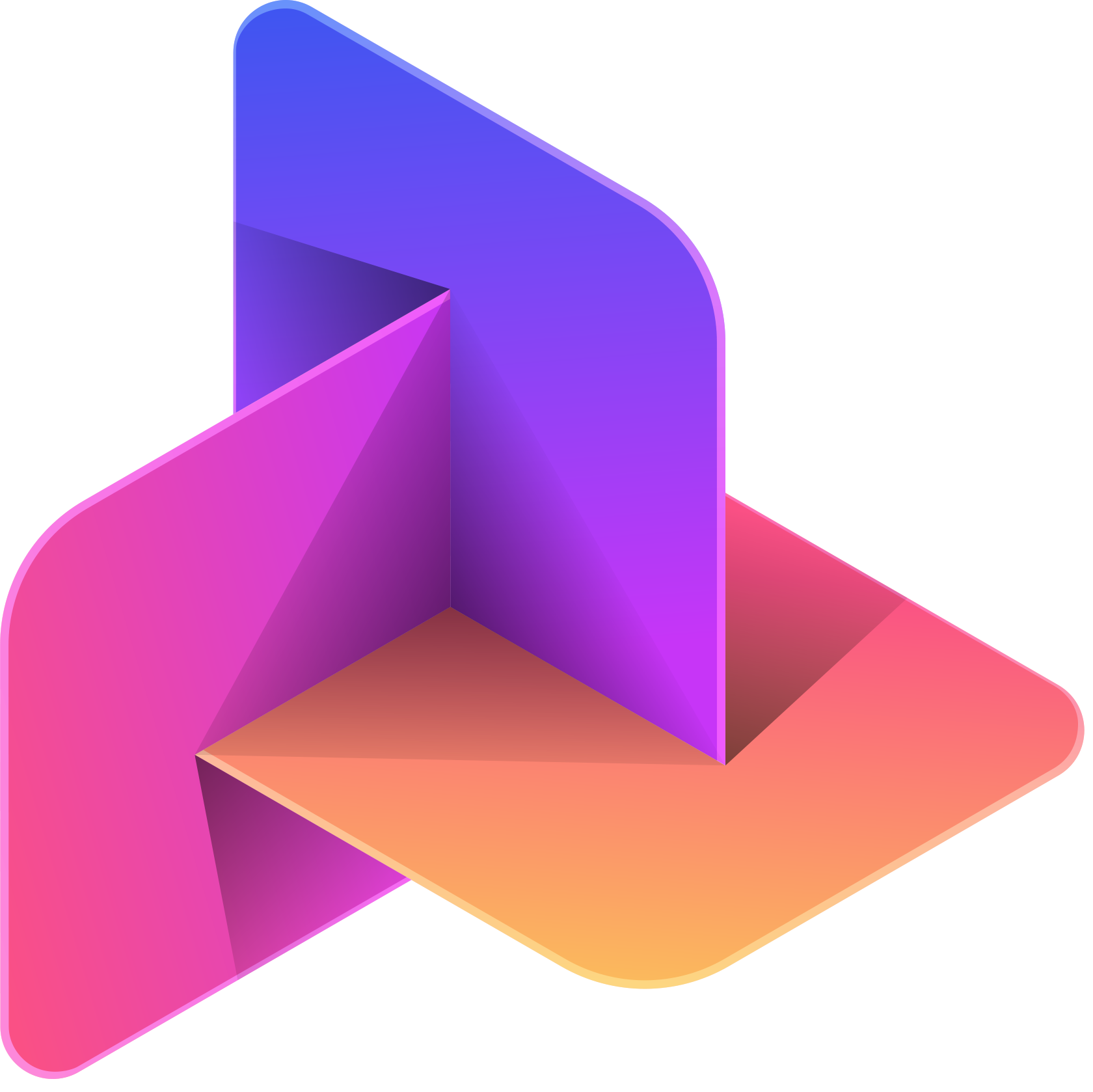}
  K-EXAONE 2.0 Technical Report \vspace{0.25em} \\ \vspace{0.5em} \large{Journey to Global Frontier-Scale Foundation Models}
}

\author{
  \comp\thanks{The complete list of authors who contributed to this work can be found in Appendix~\ref{appendix:contributors}.}
}

\begin{document}

  \maketitle

  \vfill
  \input{resources/fig_main_result}

  \input{sections/01_introduction}

  \input{sections/02_modeling}

  \input{sections/03_cpt}

  \input{sections/04_midtraining}

  \input{sections/05_posttraining}

  \input{sections/06_evaluation}

  \input{sections/07_limitation}

  \input{sections/08_deployment}

  \input{sections/09_conclusion}

  \clearpage
  \appendix
  \input{sections/99_appendix}

  \clearpage
  \bibliographystyle{plain} 
  \bibliography{refs} 

\end{document}

%% file: sections/00_abstract.tex
This technical report presents \textbf{\model}, an open-weight multilingual foundation model developed by LG AI Research as a step in our effort toward global frontier-scale foundation models.
Rather than training from scratch, we upcycle K-EXAONE and expand its architecture, yielding a Mixture-of-Experts (MoE) model with 750B total parameters and approximately 37B activated per token---more than three times the capacity of its predecessor.
\model supports context lengths of up to 256K tokens and expands multilingual coverage from six to ten languages.
Its training pipeline combines continual pre-training, difficulty-focused mid-training, and post-training to strengthen reasoning, agentic coding, multilingual capability, and safety grounded in Korean sociocultural contexts.
Across nine evaluation categories selected to reflect the conditions of practical use, \model improves over K-EXAONE and remains competitive with open-weight models, showing its largest gains in agentic coding and long-context understanding and its clearest strengths in long-context retrieval and safety.
Released under the Apache 2.0 license, \model enables the wider AI ecosystem to evaluate, deploy, adapt, and build upon it, while marking the beginning---rather than the endpoint---of our challenge toward the global frontier.

%% file: resources/fig_main_result.tex
\begin{figure}[!bh]
  \centering
  \includegraphics[width=\textwidth]{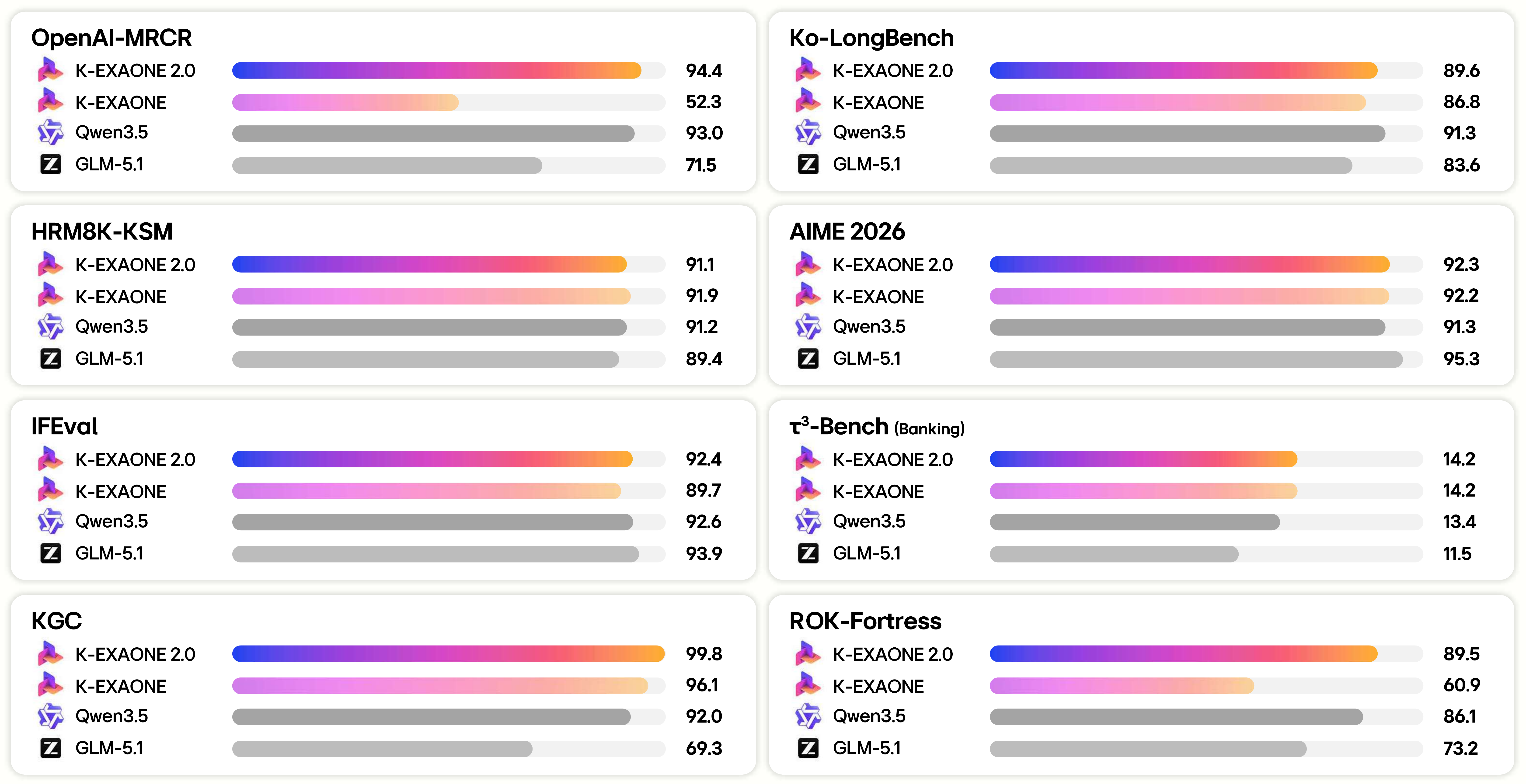}   
  \caption{The main evaluation results of K-EXAONE 2.0.}
  \label{fig:main_figure}
\end{figure}

%% file: sections/01_introduction.tex
\newpage

%
%
\section{Introduction}
\label{introduction}

The development of large language models (LLMs) has entered a period of intense global competition.
Proprietary models continue to define much of the frontier, while open-weight models have steadily narrowed the gap through advances in model scale, data, and training methodology~\citep{gptoss2025, yang2025qwen3technicalreport, llama4}.
Recent Mixture-of-Experts (MoE) models have reached hundreds of billions of total parameters and, in some cases, entered the trillion-parameter regime~\citep{deepseekai2025deepseekv3technicalreport, kimiteam2026kimik3openfrontier, deepseekv4}.
For countries seeking to establish durable AI capabilities, however, access to frontier models is not sufficient.
What matters is the domestic capacity to build, operate, evaluate, and continuously improve models grounded in their own linguistic, cultural, and institutional contexts.

For this purpose, the Korean government initiated a strategic program that provides essential computational resources, including GPUs, for the development of large-scale foundation models.
Through this program, LG AI Research developed K-EXAONE~\citep{choi2026kexaonetechnicalreport}, a MoE model with 236B total parameters and 23B parameters activated per token.
K-EXAONE demonstrated that a globally competitive foundation model supporting a 256K-token context and six languages could be developed domestically.
Building on this achievement, \model, the second-phase model developed under the same program, extends the effort toward the global frontier through a substantially larger architecture and a broader range of deployment-oriented capabilities.
This report presents the progress made in scaling, training, and evaluating the model during the current project period.

Building on K-EXAONE, we developed \model around a single objective: to advance toward a frontier-scale foundation model whose increased capacity translates into reliable capabilities under practical deployment conditions, while carrying forward the accumulated investment of the previous generation.
Training a substantially larger model from scratch would have discarded not only the parameters learned by K-EXAONE but also the computation, data, and training practices accumulated during its development.
We therefore upcycled K-EXAONE by expanding the architecture along both depth and expert capacity, stabilizing the expanded model, and resuming large-scale continual pre-training.
The resulting model contains 750B total parameters---more than three times the total parameter count of its predecessor---with approximately 37B parameters activated per token.
We retained core architectural choices that continued to serve this objective, including the hybrid attention architecture introduced in K-EXAONE~\citep{choi2026kexaonetechnicalreport,wolfetal2020transformers}, which interleaves sliding-window and global attention to support efficient long-context processing at context lengths of up to 256K tokens.
To make the resulting scale practical at inference time, \model retains a Multi-Token Prediction (MTP) module trained jointly with the backbone and adds a DSpark~\citep{dspark} drafter trained after the final model weights were fixed, providing two self-speculative decoding paths on the same target weights.

The model's increased capacity was then deliberately directed through difficulty-focused mid-training and post-training toward capabilities required in practical use.
The post-training pipeline incorporates online reinforcement learning and internally developed optimization methods, including AGAPO~\citep{bae2026exaone40unifiedlarge} and \textsc{GrouPER} (Group-wise SimPER)~\citep{choi2026kexaonetechnicalreport}, to improve instruction following, reasoning, factual reliability, and alignment.
Across mid- and post-training, we placed particular emphasis on advanced reasoning and agentic coding, where the previous model left substantial headroom, using multi-step software-engineering tasks that require the model to understand codebases, interact with tools, and maintain coherent execution over extended workflows.
We also expanded multilingual coverage from six to ten languages---Korean, English, Spanish, German, Japanese, Vietnamese, French, Italian, Polish, and Portuguese---together with their corresponding training, evaluation, and safety data.
Safety was integrated throughout model development rather than appended as a post hoc filtering step.
Through red teaming and human evaluation with experts in values and relevant domains, we refined the criteria for more than 100 existing risk areas and added 70 new ones, expanding the Korea-Augmented Universal Taxonomy and incorporating the revised criteria into both safety training and evaluation.
Together, these choices treat scale not as an end in itself, but as the foundation for capabilities that can be efficiently operated, broadly applied, and reliably deployed.

We evaluated \model across nine categories---world knowledge, mathematics, coding and agentic coding, agentic tool use, instruction following, long-context understanding, Korean language understanding, multilingual capabilities, and safety---selected to validate the capabilities required for practical deployment.
The evaluation shows consistent improvements over K-EXAONE, with the largest gains in agentic coding and long-context understanding, as well as the clearest advantages over comparable open-weight models in long-context retrieval and safety.
We release \model under the Apache 2.0 license so that the wider AI ecosystem can inspect, deploy, and build upon it.

%% file: sections/02_modeling.tex
%
%
\section{Modeling}
\label{modeling}

\model builds upon the Mixture-of-Experts (MoE) architecture of its predecessor K-EXAONE, enabling resource-efficient scaling of model capacity. 
To achieve stronger performance, the model is scaled up to 3.2× the size of its predecessor, with total parameters increasing from 236B to 750B, while the number of activated parameters grows only from 23B to 37B. 
To make this expansion effective, \model is initialized by upcycling the weights of K-EXAONE, improving training efficiency. 
It incorporates Multi-Token Prediction (MTP) and DSpark modules to enhance inference efficiency.

%
%
\subsection{Model Configurations}
\label{subsec:model_configurations}

As illustrated in Figure~\ref{fig:model_architecture}, \model is a fine-grained sparse MoE model that supports both MTP and DSpark modules, which can be applied independently.
The architecture consists of 78 layers: two dense layers followed by 76 MoE layers. For the last 16 MoE layers, we adopt Clamped SwiGLU for training and inference stability, as detailed in Section~\ref{subsec:model_upcycling}. 

\input{resources/fig_model_architecture}

As shown in Table~\ref{tab:model_config}, compared to K-EXAONE, the number of layers is increased from 48 to 78 and the number of experts per layer is doubled from 128 to 256, while dimensions of each expert remain unchanged.  

\input{resources/tab_model_config}

\model inherits several core design choices from K-EXAONE. 
A single shared expert is employed, such that one shared expert and the top-8 routed experts are activated per token during inference. 
The routing mechanism also follows the same design, employing sigmoid-based scoring, sequence-level load balancing, and a dropless routing policy. 
The hybrid attention architecture, QK Norm, and SWA-only RoPE are retained, which together enable stable training and cost-efficient long-context modeling.

On the inference side, \model additionally adopts DSpark, a drafting module, alongside the MTP module retained from K-EXAONE. The MTP module improves future-token prediction during training and serves as a self-drafting head at inference time. DSpark further accelerates decoding by generating an entire draft block in a single forward pass, achieving higher acceptance rates and greater speedup than MTP. Detailed descriptions are provided in Section~\ref{subsec:inference_optimization}.

%
%
\subsection{Model Upcycling}
\label{subsec:model_upcycling}

Growing a trained model along depth and width and continuing to train it is an established way to cut pre-training cost~\cite{stagedtraining}. 
\model is \emph{upcycled} from K-EXAONE by expanding the model along both the depth axis ($48 \rightarrow 78$ layers) and the width axis ($128 \rightarrow 256$ experts), followed by continued pre-training. 
The number of activated experts (top-8 plus one shared expert), the sliding-window size, the routing policy, and the tokenizer are inherited unchanged.

\paragraph{Depth up-scaling}
The general concept of depth up-scaling has recently been explored in various forms, including concatenating trimmed copies~\cite{solar}, repeated stacking~\cite{gstack}, and identity-initialized block insertion~\cite{llamapro}. 
However, our methodology is structurally distinct from these approaches. 
K-EXAONE repeats a four-layer unit of three local sliding-window attention layers (L) followed by one global attention layer (G), which we call an \emph{LLLG block}. 
We perform depth expansion at the block level rather than at the individual-layer level, increasing the parent model from 12 to 19 blocks by repeating blocks selected from the middle of the stack. 
In our experiments, this strategy yielded better results than repeating blocks near either the input or the output. 
Ahead of these blocks sit two dense layers, assigned global attention and a sliding window of 4096 respectively, against the 128-token window used elsewhere. 
\model therefore has a total of 78 layers.

\input{resources/fig_swiglu_clamp}

\paragraph{Width up-scaling}
Each expert is duplicated together with its router row~\citep{komatsuzaki2023sparse}.
Duplication alone leaves the two copies exactly tied --- they receive identical gradients and never differentiate, so the added capacity goes unused. 
We therefore add a \emph{random rotation noise} to the duplicated experts, which breaks the symmetry in a norm-preserving way. 
K-EXAONE inherits the DeepSeek-style router with per-expert score biases, and we observed no load-balancing pathology between original and newly added experts: the routing load stays balanced across all 256 experts with no intervention beyond the inherited bias update.

\paragraph{Training stability}
As continued training progressed, the activation magnitudes of a small number of experts in the deeper layers grew far beyond those of their peers, degrading both low-precision training and low-precision serving. 
SwiGLU can produce very large outputs even when weight decay suppresses the weight norms themselves, and does so once an expert's gate and up projections become aligned~\citep{fishman2025fp8, stepfun2026step35}. 
We therefore clamp the two SwiGLU branches element-wise before they are combined, bounding the gate branch from above and the linear branch on both sides at a threshold of 7.0 and leaving the routing decision untouched; the same construction is used in gpt-oss~\citep{gptoss2025} and DeepSeek-V4~\citep{deepseekv4}.

Figure~\ref{fig:swiglu-clamp-peak} shows the effect: with the clamp disabled at inference, the peak SwiGLU activation grows monotonically with depth and reaches $6862$ at the last layer, while the clamp holds it at $\mathrm{silu}(\tau)\cdot\tau=48.96$.

%
%
\subsection{Inference Optimization via Speculative Decoding} 
\label{subsec:inference_optimization}

\model supports speculative decoding through two drafting paths that attach to the same target weights. The Multi-Token Prediction (MTP) head is trained jointly with the backbone from pre-training onward and is refined before the RL stage to accelerate rollouts. 
Once the final weights were fixed, we additionally trained a DSpark~\citep{dspark} drafter, which is the stronger option for production serving. 

\paragraph{MTP with total-variation loss}
The MTP head is trained with the conventional cross-entropy objective alongside the backbone. 
Before the RL stage we refine it with an end-to-end total-variation loss, which optimizes the multi-step acceptance rate under rejection-sampling verification directly rather than through the cross-entropy surrogate~\citep{mtprs}.

\paragraph{DSpark drafter}
DSpark~\citep{dspark} is a semi-autoregressive drafter. 
A block-diffusion backbone~\citep{dflash} predicts the whole draft block in a single forward pass, conditioned on intermediate hidden states taken from several layers of the target and injected into the keys and values of every draft layer; a lightweight sequential module then refines the block left to right, restoring the intra-block dependencies a fully parallel drafter forgoes. 
Our drafter has five layers and a block size of 7, and is trained on data generated by the target itself.

\paragraph{Results}
We compare the two drafters with an identical draft budget ($\gamma=7$).
As shown in Table~\ref{tab:spec_main}, DSpark leads the MTP head in every cell measured, by 32$\sim$66\% in acceptance length, and end-to-end speedup rises from 1.27$\sim$1.77$\times$ to 1.81$\sim$2.57$\times$.

\input{resources/tab_spec_decoding}

%% file: resources/fig_model_architecture.tex
\begin{figure}[!htbp]
  \centering
  \includegraphics[width=0.95\textwidth]{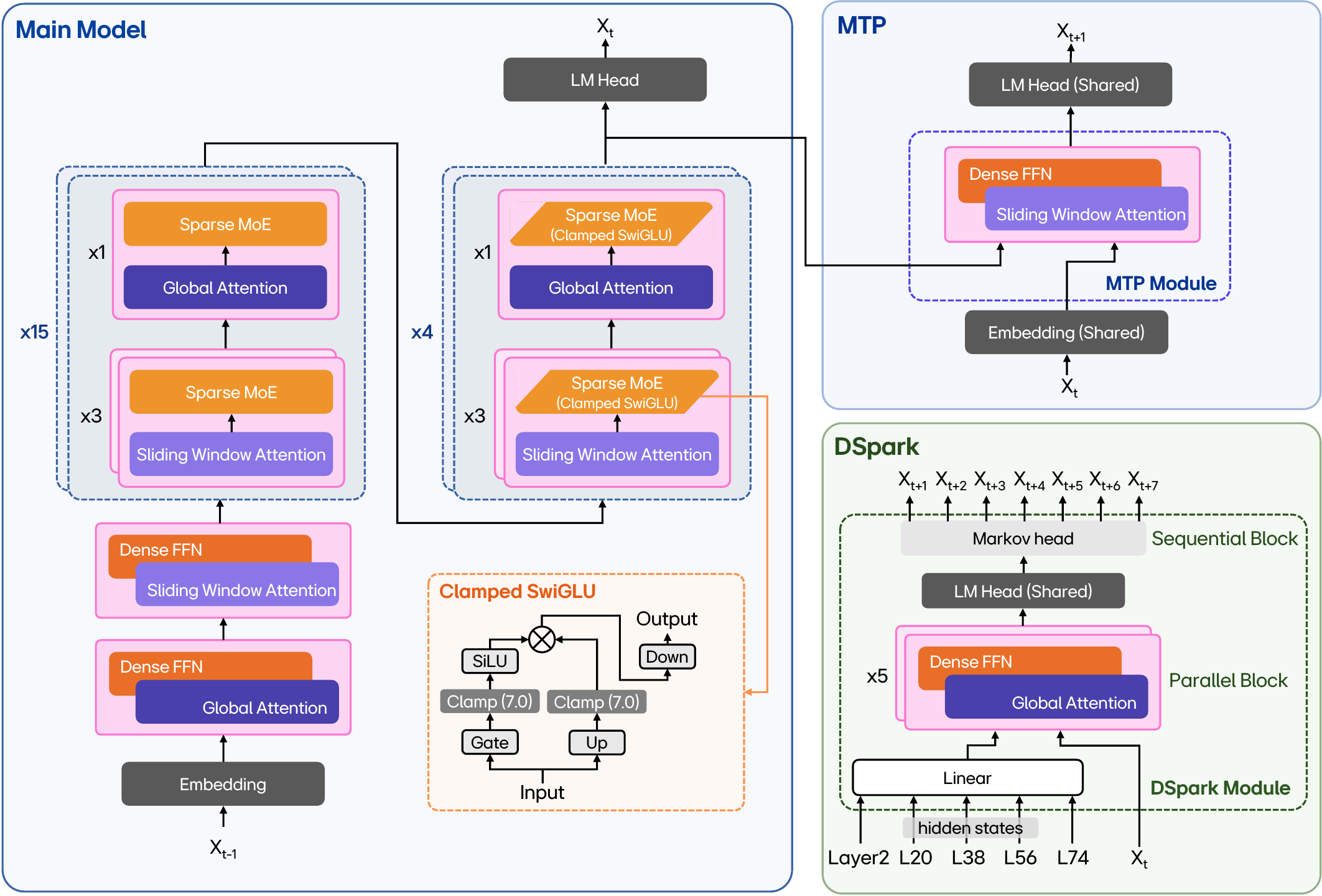}
  \caption{An illustration of \model model architecture. (Left): Main Model. (Right): MTP and DSpark modules. Incorporating either the MTP or DSpark module into the main model accelerates sequence generation. The main model consists of two initial dense layers followed by Mixture-of-Experts (MoE) layers. The sliding window attention (SWA) in the second layer employs a window size of 4096, whereas a window size of 128 is used for all other SWA layers. In the Sparse MoE layers, 8 routed experts are selected from a pool of 256 experts and deployed alongside one shared expert. To ensure stability during both training and inference, the last 16 layers of the main model apply Clamped SwiGLU with a limit value of 7.0 to the experts.}
  \label{fig:model_architecture}
\end{figure}

%% file: resources/tab_model_config.tex
\begin{table}[!htbp]
\centering
\small
\caption{Model configurations of \model and K-EXAONE.}
\label{tab:model_config}
\vspace{2mm}

\setlength{\tabcolsep}{10pt}
\begin{tabular}{l l r r}
\toprule
Block & Configuration & \model & K-EXAONE \\
\midrule
\multirow{7}{*}{Main Block}
 & Layers (Total/SWA/GA)            & 78 / 58 / 20   & 48 / 36 / 12 \\
 & Sliding Window Size              & 128            & 128 \\
 & Attention Heads (Q/KV)           & 64 / 8         & 64 / 8 \\
 & Head Dimensions                  & 128            & 128 \\
 & Experts (Total/Shared/Activated) & 256 / 1 / 8    & 128 / 1 / 8 \\
 & Experts Dimensions               & 2,048          & 2,048 \\
 & Parameters (Total/Activated)     & 750B / 37B     & 236B / 23B \\
\midrule
\multirow{4}{*}{MTP Block}
 & Layers (Total/SWA/GA)            & 1 / 1 / 0  & 1 / 0 / 1 \\
 & Attention Heads (Q/KV)           & 64 / 8  & 64 / 8 \\
 & Head Dimensions                  & 128     & 128 \\
 & Parameters                       & 0.52B   & 0.52B \\
\midrule
\multirow{4}{*}{DSpark Block}
 & Layers (Total/SWA/GA)            & 5 / 0 / 5  & -- \\
 & Attention Heads (Q/KV)           & 64 / 8  & -- \\
 & Block Size ($\gamma$)            & 7  & -- \\
 & Parameters                       & 2.53B  & -- \\
\bottomrule
\end{tabular}

\end{table}

%% file: resources/fig_swiglu_clamp.tex
\begin{wrapstuff}[type=figure, r, top=0, width=0.4\textwidth]
  \centering
  \includegraphics[width=\linewidth]{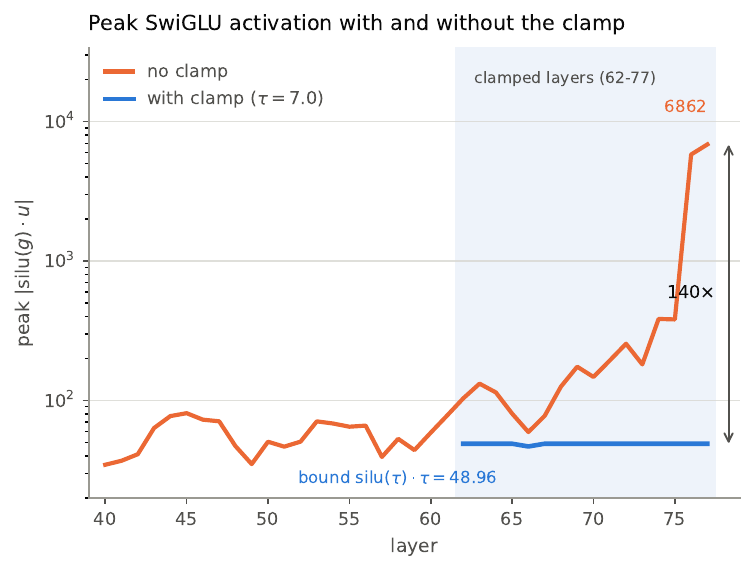}
  \captionsetup{hypcap=false}
  \caption{Peak SwiGLU activation over all routed experts, with and without the clamp ($\tau=7.0$, shaded layers 62$\sim$77).}
  \label{fig:swiglu-clamp-peak}
\end{wrapstuff}

%% file: resources/tab_spec_decoding.tex
\begin{table}[!htbp]
\centering
\small
\caption{Acceptance length and end-to-end speedup over non-speculative decoding, on the same \model (FP8) target with the same draft budget
($\gamma=7$). Each cell is \emph{non-thinking} / \emph{thinking} at temperature 1.0 on TP8, 8$\times$ H200.}
\label{tab:spec_main}

\setlength{\tabcolsep}{15pt}
\begin{tabular}{llcccc}
\toprule
\multirow{2}{*}{Domain} & \multirow{2}{*}{Benchmark} &
\multicolumn{2}{c}{Acceptance length} & \multicolumn{2}{c}{E2E speedup} \\
\cmidrule(lr){3-4} \cmidrule(lr){5-6}
 & & MTP & DSpark & MTP & DSpark \\
\midrule
\multirow{3}{*}{Math} & GSM8K     & 3.58 / 3.13 & \textbf{5.25 / 5.20} & 1.72 / 1.55 & \textbf{2.49 / 2.56} \\
                      & MATH-500  & 3.60 / 3.16 & \textbf{4.95 / 4.58} & 1.76 / 1.55 & \textbf{2.44 / 2.28} \\
                      & AIME 2026 & 3.00 / 2.73 & \textbf{4.00 / 3.60} & 1.50 / 1.36 & \textbf{2.01 / 1.81} \\
\midrule
\multirow{2}{*}{Code} & HumanEval & 3.67 / 2.61 & \textbf{5.41 / 3.81} & 1.77 / 1.30 & \textbf{2.57 / 1.92} \\
                      & MBPP      & 3.14 / 2.55 & \textbf{4.19 / 3.60} & 1.53 / 1.27 & \textbf{2.05 / 1.81} \\
\bottomrule
\end{tabular}
\end{table}

%% file: sections/03_cpt.tex
%
%
\section{Continual Pre-training}
\label{sec:continual_pre_training}

\model largely follows the pre-training data recipe established for K-EXAONE, including its overall data composition, quality filtering, and synthetic data generation methods. 
To scale up the model, we apply depth upscaling during the early stage of pre-training. 
To maintain training stability following this expansion, we first perform a healing stage using a portion of the later-stage K-EXAONE training data. 
After healing, the model is further trained on an additional 8T tokens using the \model pre-training data mixture. 
We also extend the data recipe with two new synthetic data generation methods to improve knowledge acquisition and reduce training loss. 
The first constructs synthetic data through Active Reading (AR)~\citep{lin2025learning}. 
The second extends thinking-augmented pre-training~\citep{wang2025thinkingaugmentedpretraining} to incorporate latent intermediate reasoning processes~\citep{ruan2025reasoning}.

%
%
\subsection{Knowledge-Oriented Synthetic Data}
We compare Active Reading (AR) with conventional textbook-style generation for parametric knowledge acquisition. 
Starting from the same late-stage checkpoint of K-EXAONE, we independently train three models. 
The baseline model is trained on 30B tokens from the original pre-training mixture. 
The AR and textbook-style variants are each trained on a 40B-token mixture consisting of the same 30B baseline data and an additional 10B synthetic tokens generated from Wikipedia documents using the corresponding method. 
As shown in Table~\ref{tab:ar_ablation}, AR achieves the largest average improvement over the initial checkpoint, driven primarily by its gains on ARC-C, whereas textbook-style generation performs best on MMLU and GSM8K. 
Based on these results, we apply AR selectively to knowledge-intensive domains. 
To reduce the cost of generating strategy prompts for every document, we collect effective AR prompts and reuse them across related documents, forming a lightweight pseudo-AR pipeline.

\input{resources/tab_AR}

%
%
\subsection{Korean Data}
For K-EXAONE~2.0, we focus on improving Korean-language proficiency and understanding of Korean history and culture. 
To this end, we collect high-quality Korean data from public institutions, including specialized books and challenging questions from Korea Data Industry Promotion Agency (K-DATA)\footnote{https://www.kdata.or.kr} and National Information Society Agency (NIA)\footnote{https://www.nia.or.kr}, language and cultural resources from the National Institute of Korean Language \footnote{https://www.korean.go.kr}, and historical materials from the Northeast Asian History Foundation\footnote{https://nahf.or.kr/web/portal/main}.

We evaluate institution-sourced Korean cultural and historical data, as well as open-source Korean data with high educational value. 
As shown in Table~\ref{tab:korean_data_ablation}, the former performed better on Korean culture and history benchmarks, whereas the latter achieved higher average scores on knowledge-intensive benchmarks. 
These results suggest that general knowledge transfers effectively across languages, while country-specific historical and cultural knowledge benefits substantially from high-quality local data sources.

\input{resources/tab_korean_exp}

%
%
\subsection{Multilingual Data}
In addition to the six languages supported by the original K-EXAONE, we expand language coverage by adding four new languages—French, Italian, Polish, and Portuguese—resulting in a total of ten supported languages. 
From FineWeb2~\citep{penedo2025fineweb2pipelinescale}, we construct high-quality organic datasets for each language corpus through our in-house data filtering pipeline, with the goal of enhancing fundamental language capabilities and promoting cross-lingual knowledge transfer.
We further design the Continual Pre-training dataset to support robust multi-task performance by incorporating translation data and synthetic QA datasets. 
The final set of languages supported by K-EXAONE 2.0, along with per-language performance on multilingual tasks, is provided in Appendix~\ref{appendix:multilingual}.

%
%
\subsection{Compute Scaling on Korean Data}
Figure~\ref{fig:kor_scaling} plots the loss on the Korean subset of the \model pre-training mixture against estimated cumulative training compute. 
Much of this subset consists of newly constructed internal data with limited public exposure. 
GLM-5 and EXAONE 4.0 are therefore shown only as external references and are not treated as directly comparable baselines. 
Within the K-EXAONE lineage, the loss decreases consistently from the small-scale K-EXAONE model to K-EXAONE and \model. 
Beyond increased model capacity and compute, this trend plausibly reflects the growing volume, coverage, and diversity of newly constructed Korean data across generations, with thinking-augmented and latent-thought data construction offering an additional, though not isolable, contribution.

\input{resources/fig_kor_scaling}

%% file: resources/tab_AR.tex
\begin{table}[!htbp]
\centering
\small
\captionsetup{skip=6pt}
\caption{Absolute score changes relative to the initial late-stage checkpoint of K-EXAONE. All models are trained independently from the same
checkpoint. The baseline model is trained on 30B tokens from the original pre-training mixture, while the Active Reading and textbook-style models are each trained on a 40B-token mixture consisting of the same 30B-token baseline data and an additional 10B synthetic tokens generated from Wikipedia
documents.}
\label{tab:ar_ablation}

\setlength{\tabcolsep}{10pt}
\begin{tabular}{cccccc}
\toprule
Method & ARC-C~\citep{clark2018think} & MMLU~\citep{hendrycks2020measuring} & GSM8K~\cite{cobbe2021training} & HellaSwag~\citep{zellers2019hellaswag} & Avg. \\
\midrule
Baseline Dataset & +0.00         & -0.25          & +1.51 & \textbf{+0.59}   & +0.46 \\
Active Reading   &\textbf{+1.54} & -0.11          & +1.21 & +0.34   & \textbf{+0.75} \\
Textbook-style   & -0.42         & \textbf{+0.55} & \textbf{+1.52} & -0.02   & +0.41 \\
\bottomrule
\end{tabular}

\end{table}

%% file: resources/tab_korean_exp.tex
\begin{table}[!htbp]
\centering
\small
\captionsetup{skip=6pt}
\caption{Performance comparison on Korean benchmark categories across different data sources.}
\label{tab:korean_data_ablation}

\setlength{\tabcolsep}{12pt}
\begin{tabular}{lccc}
\toprule
Data Source & Culture \& History & Knowledge \& Reasoning & Avg. \\
\midrule
Institution-sourced Data & \textbf{68.23} & 46.47 & \textbf{57.35} \\
Open-sourced Data    & 67.19 & \textbf{47.21} & 57.20 \\
\bottomrule
\end{tabular}

\end{table}

%% file: resources/fig_kor_scaling.tex
\begin{figure}[!htbp]
  \centering
  \includegraphics[width=0.7\textwidth]{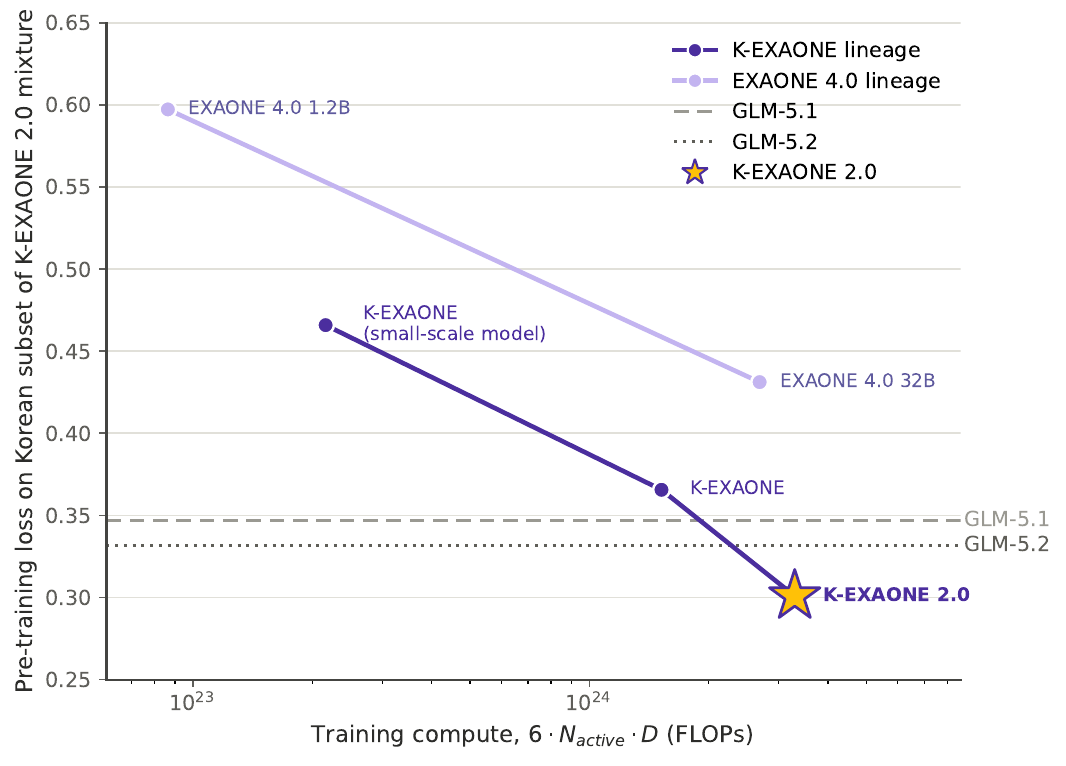}
  \caption{Pre-training loss on the Korean subset of the \model pre-training mixture versus training compute ($6 \cdot N_{active} \cdot D$, FLOPs, log scale) for the K-EXAONE and EXAONE 4.0 lineages. Dashed lines mark GLM-5.1 and GLM-5.2.}
  \label{fig:kor_scaling}
\end{figure}

%% file: sections/04_midtraining.tex
%
%
\section{Mid-training} 
\label{sec:mid_training}

We conduct mid-training in two sequential stages, progressively extending the context window from 8K to 64K and then to 256K tokens. In the first stage (Mid-Stage 1), the model is trained on 400B tokens with a 64K context window, followed by training on an additional 400B tokens with context windows exceeding 64K tokens in the second stage (Mid-Stage 2). In addition to extending the context length, mid-training is designed to improve reasoning, long-context understanding, and agentic capabilities. The longer context windows enable training on complete, untruncated long-form reasoning trajectories, cross-file dependencies in large code repositories, and multi-step tool-use workflows. To support domain adaptation, the second stage increases the proportion of data, with an emphasis on long-form reasoning, repository-level code, and agentic workflows. The following subsections describe the data construction and curation strategies for long-context adaptation, advanced knowledge and reasoning, and agentic capability development.

%
%
\subsection{Long-Context Adaptation Data}

\paragraph{Data Composition}
Long-context adaptation provides a critical foundation for robust agentic capabilities, as realistic agent workflows require models to maintain state and integrate information across extended interaction histories, multiple files, tool outputs, and intermediate reasoning steps. 
To support these capabilities, we train the model on diverse end-to-end long-context corpora comprising complete documents, large code repositories, and extended tool-use and interaction trajectories, with minimal truncation. 
We complement these corpora with synthetically constructed multi-hop data that require the model to locate, connect, and synthesize evidence distributed across distant parts of the context, thereby strengthening its ability to reason over long inputs rather than relying on localized retrieval alone. 
As the context window is extended from 64K in Mid-Stage 1 to 256K in Mid-Stage 2, we increase the proportion of these long-context examples to improve the model’s ability to maintain a coherent state, aggregate dispersed evidence, and execute complex workflows over extended horizons.

\paragraph{Long-Context Verification}
We evaluate long-context retrieval using the Needle-in-a-Haystack (NIAH)~\citep{niah} test, which measures the model’s ability to recover information inserted at varying positions within increasingly long contexts. 
As shown in Figure~\ref{fig:niah}, \model achieves perfect retrieval scores across the evaluated needle positions and context lengths of up to 256K tokens. 
This result demonstrates that the model can reliably preserve and retrieve information over contexts of up to 256K tokens, providing an important foundation for long-horizon reasoning and agentic workflows that depend on persistent access to distributed information.

\input{resources/fig_niah_template}

%
%
\subsection{Reasoning-Centric Data}

\paragraph{Filtering for Advanced Knowledge and Reasoning}
Recent reasoning tasks increasingly require knowledge and reasoning at the graduate or expert level, particularly in STEM. 
To better identify data at this level, we retrain our internal classifiers to identify corpora containing more advanced knowledge and reasoning signals in these domains. 
We also construct additional challenging knowledge data across multiple domains using an internal search agent. 
These data are introduced from Mid-Stage 2.

\paragraph{Effect of Mid-Training Stages}
We evaluate the contribution of Mid-Stage 1 and Mid-Stage 2 using the small-scale model. 
To examine whether the gains from mid-training can be preserved through post-training, we further train both the Mid-Stage 1 and Mid-Stage 2 checkpoints on the same 23B token subset of the K-EXAONE SFT dataset. 
As shown in Table~\ref{tab:mid_reasoning_ablation}, Mid-Stage 2 improves HLE~\citep{phan2025humanity} by 4.71 points over Mid-Stage 1, exceeding the 3.15-point gain obtained by applying SFT directly after Mid-Stage 1. 
Applying the same SFT data after Mid-Stage 2 yields an additional gain of 0.95 points, resulting in a total improvement of 5.66 points over Mid-Stage 1. 
These results indicate that Mid-Stage 2 not only provides a stronger foundation for advanced knowledge and reasoning, but also retains complementary benefits when followed by SFT. 
An important direction for future work is to scale the generation and curation of advanced knowledge and reasoning data, and to identify data recipes that translate mid-training improvements into larger gains after SFT.

\input{resources/tab_mid_reasoning}

%
%
\subsection{Agent Workflow Data}
For general tool use domains, we first define and categorize the core capabilities required in realistic tool use agentic scenarios, such as planning, reflection, summarization, and information aggregation. 
We then synthesize both datasets targeting each capability and agentic tool-use trajectories from expert models~\citep{tongyideepresearchteam2026tongyideepresearchtechnicalreport,hu2025stepdeepresearchtechnicalreport}. 
For coding agents, we collect and curate GitHub pull requests (PRs) across a wide range of repositories, each combining the PR description, the patch diff, and the reviewer comments, along with the relevant files and the commit history. 
These data cover diverse and realistic software development scenarios, such as bug fixing and feature implementation at the repository level.

%
%
\subsection{Tool-Calling Formats}
To avoid escaping overhead and make structured invocation more reliable, especially when tool arguments are string-heavy or contain long code spans, we adpot an XML-style tool-calling~\citep{cao2026qwen3codernexttechnicalreport,kimiteam2026kimik3openfrontier}. 
In addition, we train our model on a diverse set of tool-call templates and response formats so that it learns format-invariant tool-use behavior rather than overfitting to a single schema. 
We believe that exposure to various forms of tool-calling improves generalization across heterogeneous real-world tool environments~\citep{cao2026qwen3codernexttechnicalreport}.

%% file: resources/fig_niah_template.tex
\begin{figure}[!htbp]
  \centering
  \includegraphics[width=0.95\textwidth]{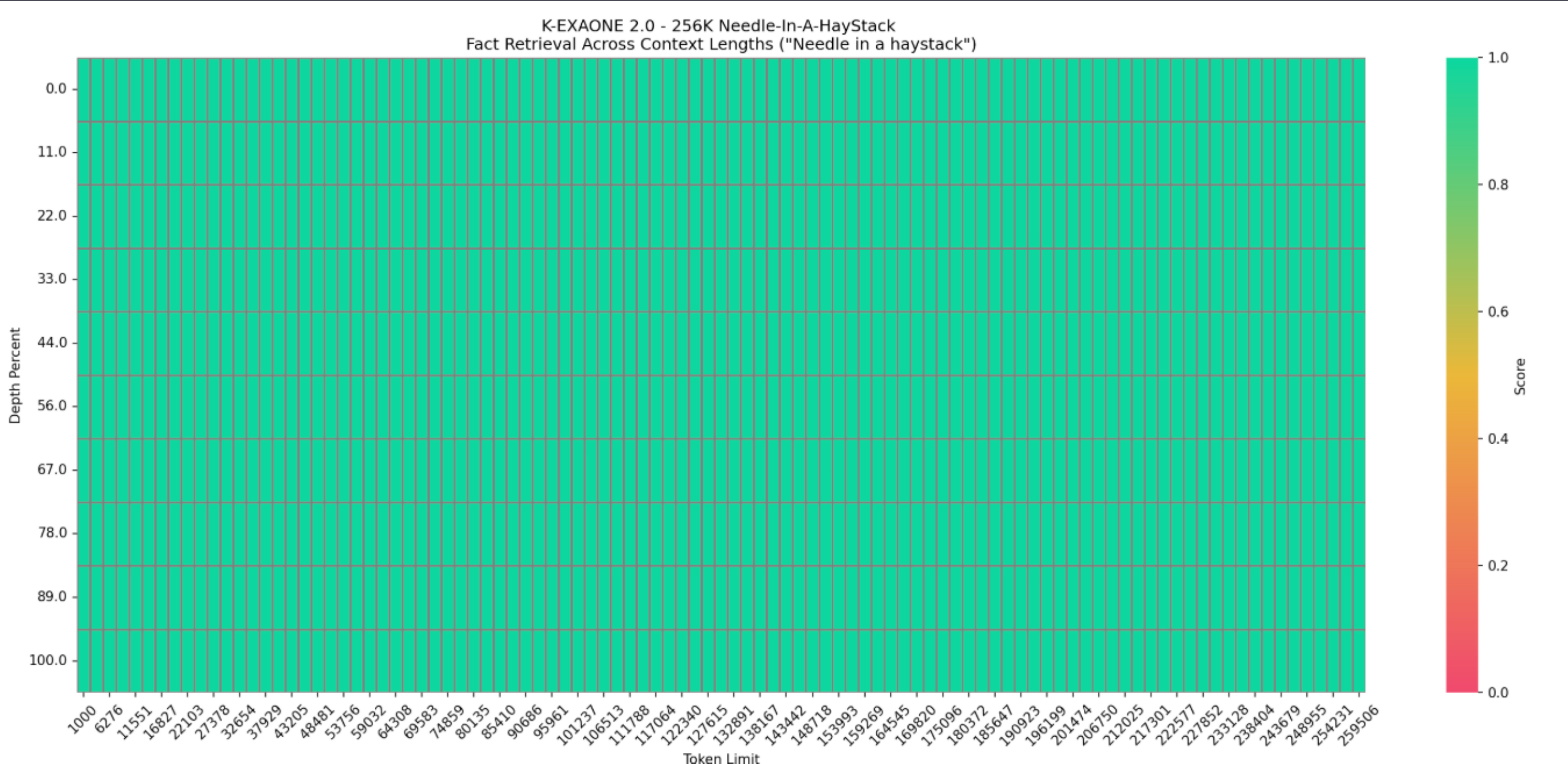}
  \caption{Needle-in-a-Haystack (NIAH) retrieval accuracy of \model across context lengths of up to 256K tokens and varying needle positions. \model maintains perfect retrieval throughout the evaluated range.}
  \label{fig:niah}    
\end{figure}

%% file: resources/tab_mid_reasoning.tex
\begin{table}[!htbp]
\centering
\small
\captionsetup{skip=6pt}
\caption{Absolute \textsc{Humanity's Last Exam} score improvements obtained from different training paths starting from the Mid Stage 1 checkpoint of the small-scale model.}
\label{tab:mid_reasoning_ablation}

\setlength{\tabcolsep}{15pt}
\begin{tabular}{lc}
\toprule
Training Stage & $\Delta$ \textsc{HLE} \\
\midrule
Mid-Stage 1 $\rightarrow$ Base SFT                           & +3.15 \\
Mid-Stage 1 $\rightarrow$ Mid-Stage 2                        & +4.71 \\
Mid-Stage 1 $\rightarrow$ Mid-Stage 2 $\rightarrow$ Base SFT & \textbf{+5.66} \\
\bottomrule
\end{tabular}

\end{table}

%% file: sections/05_posttraining.tex
%
%
\section{Post-training}
\label{sec:post_training}

Post-training comprises two stages: Supervised Fine-Tuning (Section~\ref{subsec:sft}) and Preference Learning (Section~\ref{subsec:pref}).
During Supervised Fine-Tuning (SFT), we aggregate and synthesize 350B tokens spanning the target domains of our model, including reasoning, world and expert knowledge, instruction following, and agentic systems.
By contrast, the Preference Learning stage focuses on a smaller set of targeted capabilities, including reasoning, agentic capabilities, and safety.

%
%
\subsection{Supervised Fine-Tuning}
\label{subsec:sft}

During SFT, we freeze the router parameters to preserve the expert specialization and routing patterns established during pre-training. 
We jointly train the model to support both thinking and non-thinking modes, enabling deliberate reasoning for complex tasks and direct responses for simpler interactions.
For agent training, we additionally preserve the model's thinking behavior to prevent agentic supervision from suppressing its reasoning capabilities. 
See Appendix~\ref{appendix:preserved_thinking} for further details on preserved thinking.

%
%
\subsubsection{Reasoning}

We conduct reasoning SFT to equip the model with general-purpose problem-solving capabilities across a broad range of tasks and interaction settings. 
We construct a diverse training mixture spanning mathematics, code, long-context understanding, knowledge-intensive tasks, and general conversations. 
This mixture is designed to improve the model's ability to interpret complex instructions, decompose problems into manageable steps, identify and integrate relevant information, and produce responses with an appropriate level of reasoning for each task.

\paragraph{Reasoning Data Composition}
Our reasoning data is organized into five primary domains: mathematics, code, long-context, knowledge, and general chat. 
Mathematical data emphasizes precise interpretation of problem conditions, multi-step deduction, symbolic manipulation, and logical verification. 
Code data develops procedural and algorithmic reasoning capabilities, including requirement analysis, problem decomposition, implementation, debugging, and solution refinement. 
Long-context data trains the model to identify relevant evidence distributed across lengthy inputs and to combine multiple pieces of information into a coherent solution. 
Knowledge-intensive data improves the model's ability to retrieve, select, and synthesize relevant factual knowledge, while general chat data ensures that the acquired reasoning capabilities are expressed through natural, instruction-following, and user-aligned responses. 
Together, these complementary domains support broad reasoning generalization while reducing over-specialization to individual benchmarks or problem formats.

\paragraph{Data Filtering and Quality Control}
We apply a multi-stage filtering pipeline to improve the reliability and diversity of the reasoning supervision. 
The initial rule-based filter removes malformed responses, invalid formatting, excessive repetition, incomplete generations, and samples that violate task-specific constraints. 
For selected tasks where reliable validation criteria are available, we additionally assess the validity of the answer and the consistency of the reasoning, filtering trajectories that exhibit contradictory intermediate steps, unsupported conclusions, leakage of the answer, or substantial inconsistencies between the reasoning process and the final answer. 
For a subset of tasks with objectively verifiable outputs, we further employ domain-specific validation signals, including answer-based verification for mathematical problems and execution- or test-based verification for code.

We also control artifacts introduced by synthetic data generation. 
Responses with highly templated reasoning patterns, redundant self-reflection, unnecessary restatement of the problem, or disproportionately long trajectories are removed or down-weighted. 
Since reasoning length is not necessarily correlated with reasoning quality, we prioritize trajectories that provide sufficient evidence and logical progression without excessive deliberation. 
To avoid over-representing recurring generation patterns, we perform exact and near-duplicate removal at both the prompt and response levels.

Finally, we perform decontamination against major evaluation benchmarks and related public datasets. 
Candidate training examples are compared with benchmark questions and reference answers using lexical overlap, and instances with substantial overlap are removed from the training mixture. 
This process helps reduce direct benchmark leakage and retain a cleaner set of reasoning supervision.

%
%
\subsubsection{Agent}
Although we expose the model to various tool-calling formats during mid-training, we standardize on a single tool-calling format for SFT.
We collect diverse trajectories across multiple domains of agentic systems, including general tool-calling scenarios, agentic search, and coding agents. 
In addition, our model supports a preserved thinking mode for agentic scenarios, which retains reasoning blocks across multi-turn conversations rather than discarding them at the end of each turn. 
We describe this mode in detail in Appendix~\ref{appendix:preserved_thinking}.

\paragraph{General Tool-call Scenarios}
To cover a broad range of scenarios in which users with diverse preferences and characteristics interact with agent systems, we leverage large-scale persona datasets, such as NVIDIA's Nemotron-Personas~\citep{Nemotron-Personas-USA,Nemotron-Personas-Korea}. 
For each persona, we synthesize MCP servers that the corresponding user may plausibly use, along with scenarios in which the user interacts with those servers. 
This process enables us to construct large-scale synthetic environments in which teacher models can generate trajectories, thereby allowing us to produce training trajectory data at scale.

\paragraph{Korean Public-API Tool Use}
To strengthen tool-use capabilities in Korean-specific settings, we incorporate a Korean public API tool-calling benchmark and training dataset developed by LG CNS. 
The underlying APIs span a broad range of locally relevant domains, from transportation and education to finance, law, and public administration. 
These resources complement our general tool-calling data with Korean-specific API schemas, services, and interaction patterns.

\paragraph{Agentic Search}
Following prior work on curating and synthesizing search-intensive queries for browsing and deep-research agents~\citep{xia2026open,tongyideepresearchteam2026tongyideepresearchtechnicalreport,gao2025turnsunlockinglonghorizonagentic}, we build an internal framework for generating a deep-research style query dataset. 
We organize the generated queries into two broad categories: \emph{depth-oriented} and \emph{breadth-oriented}.

Depth-oriented queries are formulated as inverted identification problems, in which explicit identifiers are replaced with indirect clues such as attributes, relations, and temporal constraints. 
This formulation aligns with BrowseComp~\citep{wei2025browsecompsimplechallengingbenchmark}-style inverted questions and subsequent synthesis pipelines based on problem inversion, obfuscation, and multi-step relational reasoning.

By contrast, breadth-oriented queries require the agent to identify a complete set of entities satisfying user-specified constraints, verify each candidate, resolve duplicates or aliases, determine when the search is sufficiently exhaustive, and present the results in a structured form, consistent with prior breadth-search benchmarks that emphasize completeness, per-item verification, de-duplication, and structured aggregation rather than the recovery of a single hidden answer~\citep{wong2026widesearch,gupta2026deepsearchqabridgingcomprehensivenessgap,huang2026wideseekadvancingwideresearch}.

We further control the mixture of these two query types and synthesize hybrid queries that combine broad candidate discovery with deep per-candidate verification. 
In internal ablation experiments, we find that balancing these query types improves the model's depth-oriented and breadth-oriented agentic-search capabilities.

\paragraph{Software Engineering (SWE)}
To construct executable and verifiable SWE tasks, we draw on existing repositories and pull requests. From repositories spanning various programming languages, including Python, Go, JavaScript, and TypeScript, we use an LLM to build Docker environments, refining each one through an iterative verification loop. 
Within the verified environments, we generate issue statements and corresponding patches from pull requests, and extract fail-to-pass (F2P) and pass-to-pass (P2P) test cases for each instance. 
Finally, we validate each task by checking that its issue statement and patch are aligned, and by confirming through an oracle run that the patch resolves the issue. 
Through this pipeline, we obtain a diverse set of tasks that mirror real-world software engineering challenges.

\paragraph{Terminal Environments}
Terminal agents operate across a wide range of workflows, such as those in software engineering, data science, and security. 
We group these workflows into task domains, and for each domain we assemble a pool of seed tasks from both curated sources and LLM synthesis~\citep{pi2026dataengineeringscalingllm, ivison2026tmaxsimplerecipeterminal}. 
These seed pools supply the background knowledge and concrete material needed to construct tasks. 
From these seeds, we use an LLM to build executable, multi-file workspace environments, together with the tasks to be solved through the terminal and the test scripts that determine whether each task has been completed correctly. 
We then repeatedly run agent rollouts on each candidate task and use the observed success rates to filter out tasks that are unsolvable or too easy. 
The rollouts also expose failure cases in the generated tasks, which we feed back into the construction stage for revision. 
Iterating this loop progressively improves the task pool and yields a validated task set suitable for training.

%
%
\subsection{Preference Learning}
\label{subsec:pref}

Following Supervised Fine-Tuning, we perform preference optimization in two sequential stages.
The first stage applies multi-task preference optimization to improve general instruction-following and reasoning capabilities across a diverse set of tasks.
The second stage applies safety-aware preference optimization to encourage appropriate abstention on unanswerable queries, refusal of harmful requests, and robustness against jailbreak attacks.
Both stages are trained using \textsc{GrouPER}~\citep{choi2026kexaonetechnicalreport}, a groupwise preference optimization objective.

\paragraph{Multi-task Preference Optimization}
The training data span reasoning domains, including mathematics, coding, and knowledge, as well as agentic and chat tasks.
For each domain, we define criteria for selecting \textit{chosen} responses and design domain-specific rewards that penalize undesirable patterns in \textit{rejected} responses.
For mathematics and coding tasks, preferred and rejected responses are primarily identified using verifiable signals.
For mathematical tasks, particularly proof-oriented problems for which exact verification is difficult, we additionally employ an LLM-as-a-judge approach to promote valid reasoning patterns and more accurate mathematical knowledge.
For chat tasks, we construct instance-specific rubrics and optimize response preferences according to the requirements of each instance.
For agentic scenarios, we evaluate not only the correctness of the agent's actions and responses, but also the quality, depth, and comprehensiveness of the final answer, which leads to more effective preference supervision.
Finally, to prevent the model from exploiting superficial cues, we select \textit{chosen} and \textit{rejected} responses with similar lengths whenever possible, improving the robustness and stability of preference optimization.

\paragraph{Safety-aware Preference Optimization}
Prompts are drawn from a combination of public and internally constructed sources and organized according to K-AUT-V2, our internally developed safety taxonomy grounded in global ethical standards and Korean-specific considerations.
The training mixture also includes queries that are unanswerable or unsupported by the provided context.
Further details on the safety criteria and data construction methodology are provided in Appendix~\ref{appendix:safety}.
For each prompt, we sample four candidate responses and score them using domain-specific reward criteria to construct the response groups used by \textsc{GrouPER}.
We determine the data mixture ratios and corresponding reward criteria through ablation studies on smaller models from the same family, as both factors substantially affect the trade-off between refusal accuracy and retention of the capabilities acquired in the preceding stage.

%
%
\subsection{Data Compliance}
\label{sec:data_compliance}

Developing AI models requires a large amount of data, and the acquisition and utilization of this data can lead to various legal issues, such as copyright infringement, intellectual property infringement, and personal information protection violations. To minimize these risks, LG AI Research conducts AI Compliance reviews throughout the entire process of data collection, AI model training, and information provision. For more detailed information, please refer to the EXAONE 3.0 Technical Report~\citep{an2026exaone3078binstruction} and the LG AI Ethics Principles~\citep{lgethics}.

%% file: sections/06_evaluation.tex
%
%
\section{Evaluation}
\label{evaluation}

%
%
\subsection{Benchmarks and Setup}
\label{subsec:benchmarks_and_setup}

We evaluate \model~on a diverse set of benchmarks spanning nine categories below:

\begin{itemize}
  \item \textbf{World Knowledge}: 
        \textsc{MMLU-Pro}~\citep{wang2024mmlupro}, 
        \textsc{GPQA-Diamond}~\citep{rein2024gpqa}, and 
      \textsc{Humanity's Last Exam}\footnote{We use the text-only subset.}~\citep{phan2025humanitysexam}
  \item \textbf{Math}: 
        \textsc{AIME 2026}~\citep{dekoninck2026matharena},
        \textsc{HMMT Feb 2026}~\citep{dekoninck2026matharena}, and
        \textsc{IMO-AnswerBench}~\citep{luong-etal-2025-towards}
  \item \textbf{Coding / Agentic Coding}: 
        \textsc{SciCode}~\citep{tian2024scicoderesearchcodingbenchmark},
        \textsc{SWE-bench Verified}~\citep{jimenez2024swebench}, and
        \textsc{Terminal-Bench 2.1}~\citep{merrill2026terminalbenchbenchmarkingagentshard}
  \item \textbf{Agentic Tool Use}: 
        \textsc{$\tau^3$-Banking}~\citep{shi2026tauknowledgeevaluatingconversationalagents} and
        \textsc{Claw-Eval}~\citep{ye2026clawevaltrustworthyevaluationautonomous}
  \item \textbf{Instruction Following}: 
        \textsc{IFEval}~\citep{zhou2023instructionfollowingevaluationlargelanguage} and
        \textsc{IFBench}~\citep{pyatkin2025generalizing}
  \item \textbf{Long Context Understanding}: 
        \textsc{AA-LCR}~\citep{artificialanalysis2025lcr}, 
        \textsc{OpenAI-MRCR}~\cite{openai2025mrcr_snapshot}, and
        \textsc{Ko-LongBench} (in-house)~\citep{choi2026kexaonetechnicalreport}
  \item \textbf{Korean}: 
        \textsc{KMMLU-Pro}~\citep{hong-etal-2025-kmmlu}, 
        \textsc{CLIcK}~\citep{kim-etal-2024-click}, and
        \textsc{HRM8K-KSM}~\citep{ko-etal-2025-understand} 
  \item \textbf{Multilinguality}\footnote{We only evaluate nine non-English supported languages on multilingual benchmarks: Korean (ko), Spanish (es), German (de), Japanese (ja), Vietnamese (vi), French (fr), Italian (it), Polish (pl), and Portuguese (pt)}: 
        \textsc{MMMLU}~\citep{hendrycks2021measuring},
        \textsc{GlobalMMLU-Lite}~\citep{singh-etal-2025-global} and
        \textsc{PolyMath}~\citep{wang2026polymath}
  \item \textbf{Safety}: 
        \textsc{KGC-Safety}\footnote{Korean Global Civic Safety Benchmark. See Appendix~\ref{appendix:safety} for details.} (in-house) and
        \textsc{ROK-Fortress}~\citep{lee2026rokfortressmeasuringeffectgeopolitical}
\end{itemize}

For baseline models, when official scores are unavailable, we evaluate them in our internal environment with inference parameters set to the recommended configuration for each model.
Please refer to Appendix~\ref{appendix:evaluation_details} for the detailed evaluation setup of each benchmark.

%
%
\subsection{Results}
\label{subsec:results}

\input{resources/tab_results_reasoning}

Table~\ref{tab:results_reasoning} presents the main evaluation results of \model in \textsc{Reasoning} mode.
Overall, \model~shows substantial improvements over K-EXAONE across agentic coding, tool use, instruction following, long-context understanding, and multilingual mathematical reasoning, while maintaining strong performance in world knowledge, mathematics, and Korean.

\paragraph{Reasoning Abilities}
On world-knowledge benchmarks, \model~achieves scores of 83.5 on \textsc{MMLU-Pro}, 82.2 on \textsc{GPQA-Diamond}, and 18.3 on the text-only subset of \textsc{Humanity's Last Exam}.
In particular, the scores on \textsc{GPQA-Diamond} and \textsc{Humanity's Last Exam} improve by 3.1 and 4.7 points over K-EXAONE, respectively.
For mathematical reasoning, \model~scores 92.3 on \textsc{AIME 2026}, 78.4 on \textsc{HMMT Feb 2026}, and 78.6 on \textsc{IMO-AnswerBench}, improving by 2.3 points on \textsc{IMO-AnswerBench}.

\paragraph{Agentic Abilities}
The model exhibits substantial gains in coding and agentic coding, achieving scores of 40.1 on \textsc{SciCode}, 68.2 on \textsc{SWE-bench Verified}, and 43.8 on \textsc{Terminal-Bench 2.1}.
Compared with K-EXAONE, the scores on \textsc{SWE-bench Verified} and \textsc{Terminal-Bench 2.1} increase by 18.8 and 13.5 points, respectively, demonstrating improved capabilities in repository-level software engineering and long-horizon terminal interaction.
For general tool use, \model achieves 80.0 on \textsc{Claw-Eval}, representing a 5.7 points improvement, while maintaining a score of 14.2 on \textsc{$\tau^3$-Banking}.

\paragraph{General Abilities}
For instruction following, \model~achieves scores of 92.4 on \textsc{IFEval} and 72.6 on \textsc{IFBench}, improving over K-EXAONE by 2.7 and 5.3 points, respectively.
The model also demonstrates substantial progress in long-context understanding.
In particular, its score on \textsc{OpenAI-MRCR} increases from 52.3 to 94.4, while its scores on \textsc{AA-LCR} and \textsc{Ko-LongBench} improve to 56.2 and 89.6, respectively.
These results indicate improved retrieval and reasoning over long inputs across both English and Korean contexts.

\paragraph{Korean and Multilingual Abilities}
Across Korean-centric benchmarks, \model~achieves scores of 69.1 on \textsc{KMMLU-Pro}, 84.2 on \textsc{CLIcK}, and 91.1 on \textsc{HRM8K-KSM}, demonstrating strong Korean professional knowledge, cultural and linguistic understanding, and mathematical reasoning.
Across the nine evaluated non-English languages, the model scores 86.6 on both \textsc{MMMLU} and \textsc{GlobalMMLU-Lite}.
It further achieves 71.3 on \textsc{PolyMath}, a 13.9 points improvement over K-EXAONE, indicating substantially enhanced multilingual mathematical reasoning.

\paragraph{Safety}
The model demonstrates strong performance on both \textsc{ROK-Fortress}, which evaluates robustness against diverse security-related prompts, and \textsc{KGC-Safety}, which assesses safety across Korean sociocultural contexts and global ethical principles. 
We continuously strengthen the model’s safety by developing an extensible framework that supports the ongoing refinement of evaluation criteria and adaptation to emerging attack patterns. Accordingly, future work should focus on progressively advancing the evaluation framework to address increasingly sophisticated safety risks.

%% file: resources/tab_results_reasoning.tex
\begin{table*}[!htbp]
\centering
\small
\caption{The main evaluation results of \model~\textsc{Reasoning} mode. Asterisk ($^*$) indicates that the scores are from each baseline model's official technical report, blog or leaderboard.}
\label{tab:results_reasoning}

\setlength{\tabcolsep}{4pt}
\begin{threeparttable}
\resizebox{1.0\textwidth}{!}{
\begin{tabular}{@{}lccccc@{}}
\toprule
\multicolumn{1}{l|}{} & \multicolumn{1}{c|}{\makecell{\textbf{~~K-EXAONE 2.0~~} \\ \smaller[1]~\textbf{(\textsc{Reasoning})}}} & \makecell{K-EXAONE \\ {\smaller[1]~(\textsc{Reasoning})}} & \makecell{Qwen3.5 \\ {\smaller[1]~(\textsc{Reasoning})}} & \makecell{GLM-5.1 \\ {\smaller[1]~(\textsc{Reasoning})}} & \makecell{DeepSeek V4 Pro \\ {\smaller[1]~(\textsc{Reasoning: max})}} \\
\midrule

\multicolumn{1}{l|}{ Architecture}       & \multicolumn{1}{c|}{MoE}  & MoE  & MoE  & MoE  & MoE \\
\multicolumn{1}{l|}{\# Total Params}     & \multicolumn{1}{c|}{750B} & 236B & 397B & 754B & 1.6T \\
\multicolumn{1}{l|}{\# Activated Params} & \multicolumn{1}{c|}{37B}  & 23B  & 17B  & 40B  & 49B \\
\midrule

\rowcolor[rgb]{0.9,0.9,0.9}\multicolumn{6}{c}{\textit{World Knowledge}} \\
\midrule

\multicolumn{1}{l|}{\textsc{MMLU-Pro}}                                         & \multicolumn{1}{c|}{83.5} & 83.8 & ~~89.8$^*$ & 86.0~~ & 87.5$^*$ \\
\multicolumn{1}{l|}{\textsc{GPQA-Diamond}}                                     & \multicolumn{1}{c|}{82.2} & 79.1 & ~~88.4$^*$ & 86.2$^*$ & 90.1$^*$ \\
\multicolumn{1}{l|}{\textsc{Humanity's Last Exam}} & \multicolumn{1}{c|}{18.3} & 13.6 & \tnote{$\dagger$}~~28.7$^*$ & 31.0$^*$ & 37.7$^*$ \\
\midrule

\rowcolor[rgb]{0.9,0.9,0.9}\multicolumn{6}{c}{\textit{Math}} \\
\midrule

\multicolumn{1}{l|}{\textsc{AIME 2026}}       & \multicolumn{1}{c|}{92.3} & 92.2 & ~~91.3$^*$ & 95.3$^*$ & 95.2~~ \\
\multicolumn{1}{l|}{\textsc{HMMT Feb 2026}}   & \multicolumn{1}{c|}{78.4} & 80.7 & ~~84.6~~ & 82.6$^*$ & 95.2$^*$ \\
\multicolumn{1}{l|}{\textsc{IMO-AnswerBench}} & \multicolumn{1}{c|}{78.6} & 76.3 & ~~80.9$^*$ & 83.8$^*$ & 89.8$^*$ \\
\midrule

\rowcolor[rgb]{0.9,0.9,0.9}\multicolumn{6}{c}{\textit{Coding / Agentic Coding}} \\
\midrule

\multicolumn{1}{l|}{\textsc{SciCode}}            & \multicolumn{1}{c|}{40.1} & 35.6 & ~~42.0$^*$ & 43.8$^*$ & 50.0$^*$ \\
\multicolumn{1}{l|}{\textsc{SWE-Bench Verified}} & \multicolumn{1}{c|}{68.2} & 49.4 & ~~76.4$^*$ & 73.6~~ & 80.6$^*$ \\
\multicolumn{1}{l|}{\textsc{Terminal-Bench 2.1}} & \multicolumn{1}{c|}{43.8} & 30.3 & ~~51.3$^*$ & 61.8$^*$ & 64.0$^*$ \\
\midrule

\rowcolor[rgb]{0.9,0.9,0.9}\multicolumn{6}{c}{\textit{Agentic Tool Use}} \\
\midrule

\multicolumn{1}{l|}{\textsc{$\tau^3$-Banking}}  & \multicolumn{1}{c|}{14.2} & 14.2 & ~~13.4$^*$ & 11.5$^*$ & 25.8$^*$ \\
\multicolumn{1}{l|}{\textsc{Claw-Eval}~{\smaller[2]~(general)}} & \multicolumn{1}{c|}{80.0} & 74.3 & ~~81.2~~ & 86.1~~ & 83.5~~ \\
\midrule

\rowcolor[rgb]{0.9,0.9,0.9}\multicolumn{6}{c}{\textit{Instruction Following}} \\
\midrule

\multicolumn{1}{l|}{\textsc{IFEval}}  & \multicolumn{1}{c|}{92.4} & 89.7 & ~~92.6$^*$ & 93.9~~ & 94.0~~ \\
\multicolumn{1}{l|}{\textsc{IFBench}} & \multicolumn{1}{c|}{72.6} & 67.3 & ~~76.5$^*$ & 76.3$^*$ & 76.5$^*$ \\ 
\midrule

\rowcolor[rgb]{0.9,0.9,0.9}\multicolumn{6}{c}{\textit{Long Context Understanding}} \\
\midrule

\multicolumn{1}{l|}{\textsc{OpenAI-MRCR}}  & \multicolumn{1}{c|}{94.4} & 52.3 & ~~93.0~~ & 71.5~~ & 92.9~~ \\
\multicolumn{1}{l|}{\textsc{AA-LCR}}       & \multicolumn{1}{c|}{56.2} & 53.5 & ~~65.7$^*$ & 62.3$^*$ & 66.3$^*$ \\
\multicolumn{1}{l|}{\textsc{Ko-LongBench}} & \multicolumn{1}{c|}{89.6} & 86.8 & ~~91.3~~ & 83.6~~ & 91.4~~ \\
\midrule

\rowcolor[rgb]{0.9,0.9,0.9}\multicolumn{6}{c}{\textit{Korean}} \\
\midrule

\multicolumn{1}{l|}{\textsc{KMMLU-Pro}} & \multicolumn{1}{c|}{69.1} & 67.3 & ~~77.4~~ & 75.8~~ & 80.5~~ \\
\multicolumn{1}{l|}{\textsc{CLIcK}}     & \multicolumn{1}{c|}{84.2} & 83.9 & ~~88.9~~ & 88.7~~ & 91.6~~ \\
\multicolumn{1}{l|}{\textsc{HRM8K-KSM}} & \multicolumn{1}{c|}{91.1} & 91.9 & ~~91.2~~ & 89.4~~ & 94.3~~ \\
\midrule

\rowcolor[rgb]{0.9,0.9,0.9}\multicolumn{6}{c}{\textit{Multilinguality}} \\
\midrule

\multicolumn{1}{l|}{\textsc{MMMLU}}           & \multicolumn{1}{c|}{86.6} & 86.2 & ~~90.6~~ & 89.7~~ & 89.6~~ \\
\multicolumn{1}{l|}{\textsc{GlobalMMLU-Lite}} & \multicolumn{1}{c|}{86.6} & 86.9 & ~~92.1~~ & 90.7~~ & 92.0~~ \\
\multicolumn{1}{l|}{\textsc{PolyMath}}        & \multicolumn{1}{c|}{71.3} & 57.4 & ~~73.3~~ & 73.8~~ & 80.9~~ \\
\midrule

\rowcolor[rgb]{0.9,0.9,0.9}\multicolumn{6}{c}{\textit{Safety}} \\
\midrule

\multicolumn{1}{l|}{\textsc{KGC-Safety}~{\smaller[2]~(in-house)}} & \multicolumn{1}{c|}{99.8} & 96.1 & ~~92.0~~ & 69.3~~ & 82.8~~ \\
\multicolumn{1}{l|}{\textsc{ROK-Fortress}}                        & \multicolumn{1}{c|}{89.5} & 60.9 & ~~86.1~~ & 73.2~~ & 47.6~~ \\
\bottomrule

\end{tabular}
}
\begin{tablenotes}\footnotesize
\item[] \textsuperscript{$\dagger$}\,Full set
\end{tablenotes}
\end{threeparttable}
\end{table*}

%% file: sections/07_limitation.tex
%
%
\section{Limitations}
\label{limitations}
 
\model language models, like all existing language models, have certain limitations and may occasionally generate inappropriate responses. 
The language model generates responses based on the output probability of tokens, and it is determined during learning from training data.
While we make every effort to exclude personal, harmful, and biased information from the training data, some problematic content may still be included, potentially leading to undesirable responses. 
Please note that the text generated by \model language models does not reflect the views of LG AI Research.

\begin{itemize}
  \item Inappropriate answers may be generated, which contain personal, harmful or other inappropriate information.
  \item Biased responses may be generated, which are associated with age, gender, race, and so on.
  \item The generated responses rely heavily on statistics from the training data, which can result in the generation of semantically or syntactically incorrect sentences.
  \item Since the models do not reflect the latest information, the responses may be false or contradictory.
\end{itemize}
	
LG AI Research strives to reduce potential risks that may arise from \model language models. 
Users are not allowed to engage in any malicious activities (e.g., keying in illegal information) that may induce the creation of inappropriate outputs violating LG AI's ethical principles when using \model language models.

%% file: sections/08_deployment.tex
%
%
\section{Deployment}
\label{deployment}

Section~\ref{appendix:license} in the Appendix provides license information for using the \model models. Understanding the license information is essential for the legal utilization of the language model.

%% file: sections/09_conclusion.tex
%
%
\section{Conclusion}
\label{conclusion}

In this report, we presented \textbf{K-EXAONE 2.0} as a starting point for realizing the potential of frontier-scale foundation models.
By upcycling K-EXAONE and expanding its depth and expert capacity, we scaled the model to 750B total parameters with approximately 37B parameters activated per token, while preserving and further developing the capabilities acquired by its predecessor.
The model retains a hybrid attention architecture supporting context lengths of up to 256K tokens and provides two self-speculative decoding paths on the same target weights: a Multi-Token Prediction (MTP) module trained jointly with the backbone and a DSpark drafter trained after the final model weights were fixed.
Together, these choices demonstrate a practical approach to increasing model capacity across generations while making the resulting scale usable at inference time.

We directed this increased capacity through continual pre-training, difficulty-focused mid-training, and post-training toward capabilities required under practical deployment conditions.
Continual pre-training stabilized the expanded architecture and extended the model's knowledge, Korean-language proficiency, and multilingual foundation.
Mid-training progressively extended the context length while concentrating on advanced reasoning, long-context understanding, repository-level coding, and multi-step tool-use workflows.
Post-training further refined instruction following, reasoning, factual reliability, agentic behavior, and alignment through supervised fine-tuning, online reinforcement learning, and preference learning.
Across these stages, we expanded multilingual coverage from six to ten languages and integrated safety throughout training and evaluation, with criteria grounded in Korean sociocultural contexts.
Together, these efforts translated architectural scale into capabilities that can be efficiently operated, broadly applied, and reliably deployed.

Evaluation on 24 benchmarks spanning nine categories selected to reflect practical deployment conditions---including world knowledge, agentic coding, long-context understanding, multilingual capabilities, safety, \textit{etc.}---shows an average improvement of more than 10\% over K-EXAONE. 
The largest gains over its predecessor were observed in coding and agentic coding, with improvements of approximately 30\% on three benchmarks, as well as long-context understanding; all of these areas were key priorities in the development of \model. 
Beyond these achievements over its predecessor, \model showed its clearest advantages over comparable open-weight models in long-context retrieval and safety.
Taken together, these results show that the model's increased capacity translated into effective behavior across a diverse range of practical settings, rather than improvements confined to a narrow set of benchmarks.
In this respect, \model marks the point at which we begin to realize the broader potential of frontier-scale foundation models in practical deployment.
At the same time, \model does not lead on every benchmark, and closing the remaining gaps will require continued advances in model scale, data, training methodology, inference efficiency, safety, and evaluation.

We release \model under the Apache 2.0 license so that the wider AI ecosystem can independently evaluate, deploy, adapt, and build upon it.
This release also enables the capabilities and limitations reported here to be examined across a broader range of applications and deployment environments.
Building on the technical and operational capabilities established through this project, our next phase targets models at and beyond the trillion-parameter scale.
\model therefore marks not the endpoint of the K-EXAONE model lineage, but the beginning of our full-scale challenge toward global frontier models.

%% file: sections/99_appendix.tex
%
%
\section{Contributors}
\label{appendix:contributors}

All authors are listed in alphabetical order by last name.

\paragraph{Core Contributors}
Eunbi~Choi, Kibong~Choi, Sehyun~Chun, Seokhee~Hong, Junwon~Hwang, Hyojin~Jeon, Ahra~Jo, Hyunjik~Jo, Yeonsik~Jo, Minhyeok~Jung, Doyoung~Kim, Heegyu~Kim, Joonkee~Kim, Seonghwan~Kim, Soyeon~Kim, Sunkyoung~Kim, Yireun~Kim, Yongil~Kim, Byungoh~Ko, Changhun~Lee, Dohaeng~Lee, Haeju~Lee, Jinsik~Lee, Kyungmin~Lee, Minwoo~Lee, Wonkee~Lee, Sangha~Park, Sungjune~Park, Kwangrok~Ryoo, Kijung~Seo, Minju~Seo, Yongwoo~Song, Sejong~Yang, Heuiyeen~Yeen

\paragraph{Contributors}
Stanley~Jungkyu~Choi, Yemuk~Choi, Yongchan~Chun, Jiwon~Ham, Dasol~Hong, Sujeong~Im, Kijeong~Jeon, Gerrard~Jeongwon~Jo, Hyeongjun~Jo, Yujin~Jo, Jiyeon~Jung, Naeun~Kang, Daeseong~Kim, Euisoon~Kim, Hayeon~Kim, Hyosang~Kim, Myoungshin~Kim, Unsol~Kim, Youchul~Kim, Chaeeun~Lee, ChaeYoon~Lee, Edward~Hwayoung~Lee, Honglak~Lee, Hwansoo~Lee, Minkyung~Lee, Sangeun~Lee, Solji~Lim, Woohyung~Lim, Chanwoo~Moon, Jueun~Mun, Jimin~Park, Seojeong~Park, Yongmin~Park, Hyerin~Seo, Donghyeon~Shin, Donghyun~Son, Eunyong~Son, Kaehyun~Um, Sihoon~Yang, Chang~En~Yea, Sihyuk~Yi, Kyungjae~Yoo, Chansik~Yoon

%
%
\clearpage
\section{Model License}
\label{appendix:license}

\begin{spacing}{1.25}
\textbf{Apache 2.0} \\
\\
Copyright © 2026 LG AI Research \\
\\
Licensed under the Apache License, Version 2.0 (the "License"); \\
you may not use this file except in compliance with the License. \\
You may obtain a copy of the License at \\
\\
http://www.apache.org/licenses/LICENSE-2.0 \\
\\
Unless required by applicable law or agreed to in writing, software \\
distributed under the License is distributed on an "AS IS" BASIS, \\
WITHOUT WARRANTIES OR CONDITIONS OF ANY KIND, either express or implied. \\
See the License for the specific language governing permissions and \\
limitations under the License.
\end{spacing}

%
%
\clearpage
\section{Evaluation Setup Details}
\label{appendix:evaluation_details}

When evaluating models, we try to follow the official evaluation setup for each benchmark. 
Following is the specific setting we use in our internal evaluation environment. Not mentioned benchmarks are evaluated under official setup.

\paragraph{Claw-Eval}
We evaluate on Claw-Eval at latest commit d3f02d4. 
A number of tasks omit the \texttt{mock\_today} field, which pins the reference date used by date-dependent graders; we patched these tasks to supply it, so that grading is deterministic across runs.

\paragraph{Terminal-Bench 2.1}
We use \textit{Terminus 2} as the default agent. 
We sample with temperature 1.0 and top-p 0.95, and allow up to 32,768 generated tokens per step within a 256K context window. 
Each task runs in a container provisioned with 8 CPU cores, 16GB of memory, and 20GB of disk, with a 4-hour timeout.

\paragraph{SWE-Bench Verified}
We use \textit{mini-SWE-agent}~\cite{yang2024sweagent} as the default agent. 
We sample with temperature 1.0 and top-p 0.95, and allow up to 32,768 generated tokens per step within a 256K context window. 
Each instance runs in a container provisioned with 4 CPU cores, 8GB of memory, and 10GB of disk, with a 4-hour timeout.

\paragraph{OpenAI-MRCR}
We follow the official \textsc{OpenAI-MRCR} protocol, requiring the model to prepend the provided alphanumeric hash. 
Scores are computed using the \texttt{difflib.SequenceMatcher} ratio. 
For each context-length bin, we average the scores from the 2-needle, 4-needle, and 8-needle settings to obtain a bin-level score. 
We evaluate bins up to 128K tokens (despite MRCR supporting contexts up to 1M) and report the macro-average over the resulting bin-level scores.

\paragraph{ROK-Fortress}
We adopt the multi-model judging framework used for TRS computation in the \textsc{ROK-Fortress}~\citep{lee2026rokfortressmeasuringeffectgeopolitical}. 
To improve evaluation reliability using more recent frontier models, we employ Kimi K2 Thinking, o4-mini, and GPT-5.2 as the judge model ensemble. 
The benchmark currently consists of four categories, which differ as described below.

\begin{itemize} 
  \item $\mathrm{V}_{\mathrm{En,US}}$: Original English with US context. 
  \item $\mathrm{V}_{\mathrm{Ko,US}}$: Korean translation with the US context preserved. 
  \item $\mathrm{V}_{\mathrm{En,KR}}$: English with Korean cultural adaptation. 
  \item $\mathrm{V}_{\mathrm{Ko,KR}}$: Korean with Korean cultural adaptation (full transcreation). 
\end{itemize}

The final score is calculated by averaging the TRS values across four  categories and subtracting the average from 100:

\begin{equation}
  \mathrm{Final\ Score} = 100 - \frac{ \mathrm{TRS}_{\mathrm{V}_{En,US}} + 
                                       \mathrm{TRS}_{\mathrm{V}_{Ko,US}} +                                       \mathrm{TRS}_{\mathrm{V}_{En,KR}} +
                                       \mathrm{TRS}_{\mathrm{V}_{Ko,K}} }
                                     {4}
\end{equation}

This inversion transforms the original risk-oriented TRS into a safety-oriented score, such that a higher score indicates better safety performance.

%
%
\clearpage
\section{Preserved Thinking}
\label{appendix:preserved_thinking}

\begin{figure}[!ht]
\centering
\includegraphics[width=\textwidth]{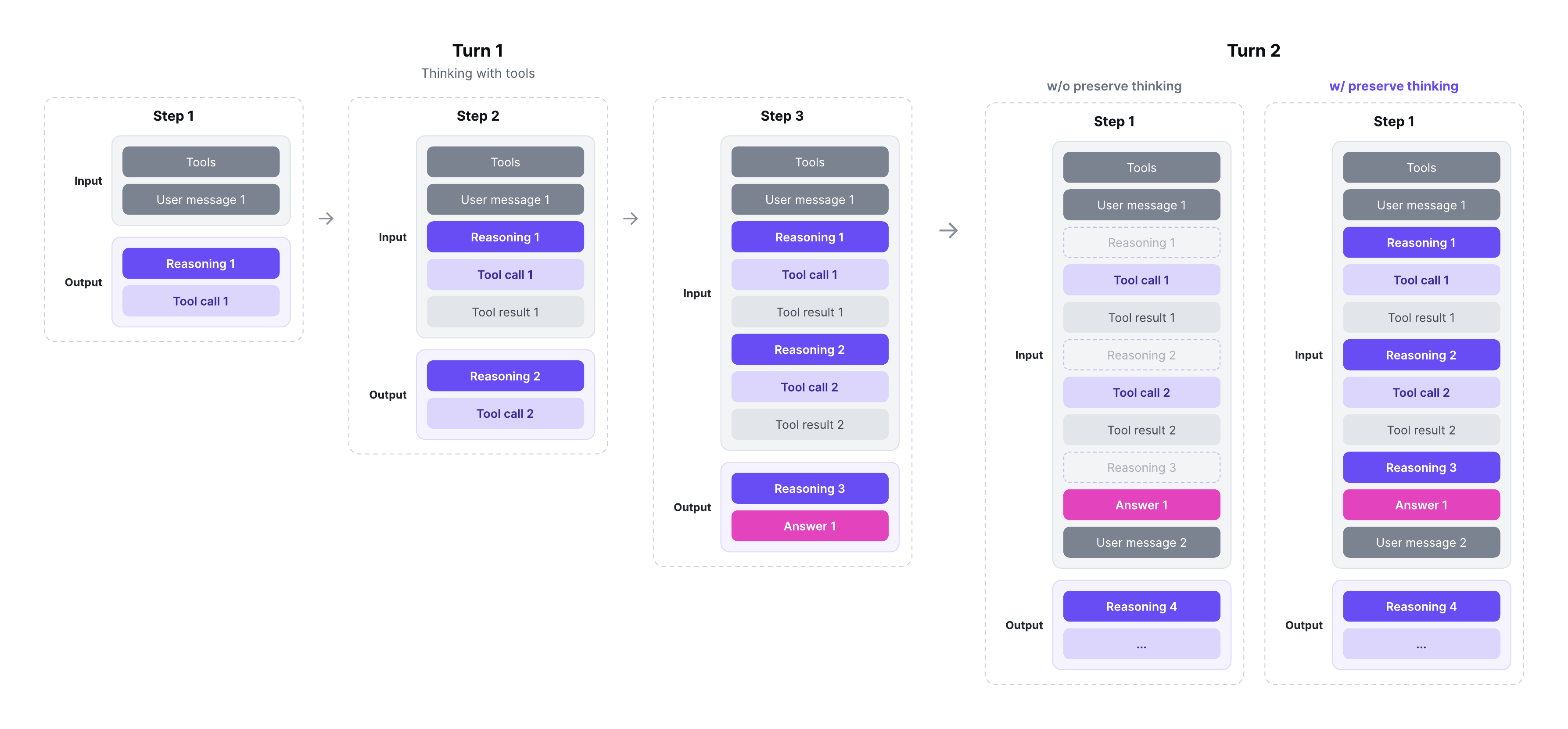}
\caption{Illustration of preserved thinking.}
\label{fig:k2_preserve_thinking}
\end{figure}

In agentic scenarios, our model supports a preserved thinking mode, which retains every reasoning block from previous turns throughout the entire conversation. 
Within a single turn, the model interleaves reasoning with tool calls, so each reasoning block is conditioned on the tool results observed so far. 
Once the turn produces its final answer, these reasoning blocks are typically dropped from the context, and the model has to re-derive its intermediate conclusions at the start of the next turn. 
Preserved thinking instead carries the full reasoning trace across the turn boundary. 
Figure~\ref{fig:k2_preserve_thinking} contrasts the two settings, where the dashed slots mark the reasoning blocks that are discarded without preserved thinking. 
This allows the model to reason consistently across turns and improves performance on difficult or long-horizon agentic tasks.

%
%
\clearpage
\section{Multilingual}
\label{appendix:multilingual}

K-EXAONE supported six languages in total: Korean, English, Spanish, German, Japanese, and Vietnamese. Starting from K-EXAONE~2.0, four additional languages—French, Italian, Polish, and Portuguese—were added, bringing the total number of supported languages to ten. \model achieves higher performance in average comparable to K-EXAONE in Table~\ref{tab:results_reasoning}. As shown in~\ref{tab:multi_langwise_scores_polymath}, \ref{tab:multi_langwise_scores_global_mmmlu}, performance gains are evenly distributed across languages, resulting in balanced multilingual capability without pronounced degradation or dominance in any single language. $\Delta_{\text{lang}}$ denotes the performance variation across languages.

Additionally, We report translation performance on WMT24++~\citep{deutsch-etal-2025-wmt24} for English and nine additional languages in both translation directions: English-to-target language and target language-to-English. Detailed results are presented in Table~\ref{tab:multi_langwise_scores_wmt24pp}.
\input{resources/tab_multilingual_langwise_results}

%
%
\clearpage
\section{Safety} 
\label{appendix:safety}

Building a sovereign model requires owning the definition of risk itself. 
Without a standard of its own, safety work inevitably falls back on taxonomies developed primarily within Western contexts or on criteria that vary from case to case.

For this reason, our previous release, K-EXAONE, introduces the Korea-Augmented Universal Taxonomy (K-AUT), which organizes potential harms into four domains and 226 detailed risk areas.
From its inception, K-AUT is designed not to become a “Galapagos” standard meaningful only within Korea.
It therefore takes universal human values, including UN declarations and internationally recognized human rights norms, as its foundation, while extending them to reflect sensitivities arising from Korea’s historical, geopolitical, and cultural context.
It also reaches beyond present-day societal risks to encompass foreseeable future risks.
Its methodology and principal categories are disclosed to domestic and international institutions through our \href{https://www.lgresearch.ai/data/cdn/upload/2025_LG%20Accountability%20Report%20on%20AI%20Ethics_Eng.pdf}{2025 Accountability Report on AI Ethics}.

In parallel with the model’s expanded multilingual capabilities, we also broaden the scope of safety training. 
In addition to the languages already supported, we incorporate four additional languages: French, Italian, Polish, and Portuguese. 
We further expand the technical coverage of our safety work by incorporating defenses against attack methods reported in prior research and extending adversarial testing to multi-turn interactions.

Yet even with this broader linguistic coverage and increasingly sophisticated attack techniques, one structural limitation remains. 
Red teaming based on established attack methods primarily tests whether risks that have already been identified and enumerated can be elicited from the model.
It is effective in assessing the coverage of known risks but comparatively weak in surfacing risks that no one has previously considered or defined. 
We therefore conclude that the remaining blind spot stems less from insufficient technical sophistication than from the limited range of perspectives involved in risk discovery.

Recognizing that safety evaluation and preparedness must evolve continuously, we seek to move beyond standards defined solely at the global or national level. 
We expand the scope of risk discovery to include field experts with direct experience in specialized professional domains. 
Through this process, we aim to identify risks that may not be visible through general-purpose safety frameworks alone and to make the model safer not only against broad societal harms but also within domain-specific contexts. 
The following sections describe this process in detail.

%
%
\subsection {Expanding the Locus of Risk Discovery}

We therefore extend the scope of risk discovery from technical experts to value and field experts. 
In partnership with the UNESCO Asia-Pacific Centre of Education for International Understanding (APCEIU), we establish and operate a Safety Teacher Advisory Council comprising 46 teachers who have completed UNESCO Global Citizenship Education (GCED)\footnote{The foundations of Global Citizenship Education (GCED) lie in UNESCO’s longstanding work on peace, human rights, and international understanding. GCED aims to empower learners of all ages to take active roles, both locally and globally, in building more peaceful, tolerant, inclusive, and secure societies. It encompasses three domains of learning: cognitive, socio-emotional, and behavioural.\href{ https://unesdoc.unesco.org/ark:/48223/pf0000227729}{(UNESCO, 2014)}} training and currently teach it in their own classrooms.

This group is particularly well suited to the task. 
Its members combine a trained sensitivity to universal values, including human rights, diversity, and inclusion, with a deep familiarity with the Korean context. 
Throughout their careers, they teach students not merely what is right or wrong as a matter of fact, but how to make such judgments in light of both universal values and the Korean context.

Recruitment is announced to several thousand GCED-certified teachers, and more than four hundred apply. 
We directly conduct the final selection, balancing six dimensions, including gender, age, school level, teaching experience, subject area, and region, to maximize the diversity of perspectives among applicants who meet the expertise requirements. 
The resulting composition is shown in Figure~\ref{fig:s_p}. 

\input{resources/fig_safety_static}

This composition targets two objectives at once.
The first is maturity of judgment: with 35 members (76\%) having more than ten years of classroom experience, the review draws on accumulated practice rather than impression.
The second is breadth of perspective. 
School levels are distributed evenly across elementary, middle, and high schools because the nature of a risk changes with a student’s developmental stage even when the model’s response remains identical.
Subject areas center on disciplines that address norms and context, including Korean, English, history, and social studies, while also encompassing mathematics, computing, arts, and special education so that the review does not converge on a single disciplinary lens. 
The purpose of this balance is ultimately to ensure representativeness in risk discovery, since which risks become visible depends heavily on who is conducting the review.

%
%
\subsection {Consultation Process with the Safety Teacher Advisory Council}

The program operates as a four-week closed loop. 
The council directly red-teams the model and proposes revisions to existing judgment criteria as well as new risk areas. 
Our research team curates these proposals and returns the resulting criteria to the council for a second round of human review. 
The adopted changes are incorporated into K-AUT and applied to model training and evaluation.

%
%
\subsection {\textsc{K-AUT-V2} : Expanded Korea-Augmented Universal Taxonomy}
\label{appendix:kgc_detail}

The council refines the judgment criteria for more than one hundred existing risk areas and identifies 70 new ones, expanding the taxonomy from 226 to 296 risk areas. Detailed statistics are provided in Table~\ref{tab:taxonomy_expansion}.

\input{resources/tab_safety_criteria}

Additionally, Figure~\ref{fig:kgc_example} presents an actual case in which the judgment criteria for reverse discrimination in a multicultural context are revised through this type-identification process.

\input{resources/fig_safety_kgc_example}

Beyond these revisions, V2 strengthens the protection of structurally vulnerable parties, including minors, subordinate parties in asymmetric relationships, and victims, as a core judgment criterion, thereby reinforcing the global citizenship dimension of the taxonomy. 
It expands coverage of Korea-specific geopolitical and historical risks, such as the North Korean nuclear issue, constitutional order, historical revisionism, and diaspora identity, as well as frontier risks, including AI goal misalignment and the erosion of human relationships caused by dependence on AI. 
It also introduces new categories covering risks that accumulate over multi-turn conversations and risks arising from the interaction itself, such as model sycophancy, misplaced empathy, and complicity with harmful premises.

These changes reflect the intended role of K-AUT: not merely as a list of prohibitions, but as a comprehensive behavioral policy that specifies what the model should refuse, how much assistance it may provide, and how refusals should be delivered.

%
%
\subsection {Making the Judgments Trustworthy}

However well suited the experts may be, the results are only as trustworthy as the procedures through which their judgments are collected and resolved. 
We therefore manage trustworthiness in three stages: whom we ask through expert selection, how we resolve divergent opinions, and whether the resulting decisions lead to actual changes in the model through verification.

\paragraph{Selection}
Partnering with APCEIU allows us to draw from a pre-qualified pool rather than recruiting experts on an ad hoc basis.
We directly apply the balancing criteria described above during the final selection process rather than delegating this responsibility.

\paragraph{Handling disagreement}
We treat disagreement as signal rather than noise.
When teachers diverge in their assessments of a given response, that divergence itself serves as evidence that the issue is genuinely sensitive, which is precisely what we seek to identify.
We therefore neither average their judgments nor discard minority views. Each response is reviewed by multiple teachers, and when their assessments diverge, we examine the specific revisions proposed by each teacher before making the final determination ourselves. 
In other words, the teachers provide the grounds for determining what is risky and why, while we decide which of those grounds becomes the final criterion and retain responsibility for that decision.
 
\paragraph{Verification}
Revised criteria have limited value unless they lead to changes in model behavior. 
We construct training data based on the revised criteria and re-evaluate the model to confirm that the intended behavioral changes occur, combining human and model-based evaluation. 
The two approaches are complementary: human evaluation reflects genuine preferences more faithfully but is slow and costly, whereas model-based evaluation provides speed and scalability while risking the transfer of the judge model’s own biases into the data.
Because such errors can propagate downstream, we reserve human evaluation for safety-critical and contested areas, use model-based evaluation where broad coverage is required, and treat human judgment as the anchor against which the judge model is continuously validated. 
Human evaluation is conducted by the AI ethics staff responsible for designing K-AUT. 
Having the authors of the criteria apply them directly minimizes the gap between how a criterion is written and how it is adjudicated in practice.

\paragraph{Conclusion and Analysis}

As shown in Table~\ref{tab:kgc_results}, the iterative refinement of evaluation criteria help mitigate model safety vulnerabilities. 
As a result, \model achieve consistently higher \textit{Safe Rates} across all evaluated dimensions of \textsc{KGC-Safety} than previous K-EXAONE models. 
These results suggest that K-EXAONE is progressively establishing a robust framework for ensuring social safety and enhancing reliability in the Korean context, thereby advancing its role as a sovereign AI model.

\input{resources/tab_safety_kgc_result}

%% file: resources/tab_multilingual_langwise_results.tex
\begin{table*}[!htbp]
\centering
\small
\renewcommand{\arraystretch}{1.2}
\caption{Multilingual performance comparison on \textsc{POLYMATH}.}
\label{tab:multi_langwise_scores_polymath}

\setlength{\tabcolsep}{10pt}
\begin{tabular}{l|cccccccc}
\toprule
 & KO & DE & ES & JA & VI & PT & FR & IT \\

\midrule
K-EXAONE        & 55.5 & 59.3 & 57.8 & 58.2 & 56.9 & 57.3 & 58.7 & 55.2   \\
\textbf{\model} & 68.8 & 70.3 & 70.4 & 73.6 & 69.5 & 71.2 & 74.0 & 72.6  \\

\bottomrule
\end{tabular}

\end{table*}

\begin{table*}[!htbp]
\centering
\small
\renewcommand{\arraystretch}{1.2}

\caption{Multilingual performance comparison on \textsc{GlobalMMLU-Lite.}}
\label{tab:multi_langwise_scores_global_mmmlu}
\setlength{\tabcolsep}{10pt}
\begin{tabular}{l|ccccccccc}
\toprule
 & KO & DE & ES & JA & VI & PT & FR & IT & PL \\

\midrule

K-EXAONE        & 86.3 & 86.5 & 88.5 & 88.0 & 84.8 & 86.5 & 88.8 & 86.3 & 86.3  \\
\textbf{\model} & 86.5 & 88.3 & 87.8 & 87.3 & 83.0 & 88.0 & 87.8 & 86.3 & 84.5  \\
\bottomrule
\end{tabular}
\end{table*}

\begin{table*}[!htbp]
\centering
\small
\renewcommand{\arraystretch}{1.2}

\caption{Multilingual performance comparison on \textsc{WMT24++}.}
\label{tab:multi_langwise_scores_wmt24pp}

\setlength{\tabcolsep}{10pt}
\begin{adjustbox}{max width=\textwidth}
\begin{tabular}{l|cccccccccc}
\toprule
 & KO & DE & ES & JA & VI & PT & FR & IT & PL & $\Delta_{\text{lang}}$  \\
\midrule
\multicolumn{11}{l}{K-EXAONE} \\
\midrule
EN $\rightarrow$ XX & 89.4 & 87.3 & 88.6 & 82.7 & 89.5 & 85.3 & 84.4 & 82.1 & 73.9 & $\pm$4.7 \\
XX $\rightarrow$ EN & 94.0 & 94.7 & 94.8 & 92.3 & 92.6 & 94.3 & 93.3 & 93.7 & 90.0 & $\pm$1.4 \\
\midrule
\multicolumn{11}{l}{\textbf{\model}} \\
\midrule
EN $\rightarrow$ XX & 87.9 & 85.1 & 87.5 & 84.4 & 87.6 & 86.5 & 83.7 & 83.6 & 78.4 & $\pm$2.8  \\
XX $\rightarrow$ EN & 93.2 & 94.4 & 94.2 & 90.7 & 91.9 & 93.2 & 92.2 & 92.7 & 89.8 & $\pm$1.4  \\
\bottomrule
\end{tabular}
\end{adjustbox}

\end{table*}

%% file: resources/fig_safety_static.tex
\begin{figure}[!htbp]
  \centering
  \includegraphics[width=\textwidth]{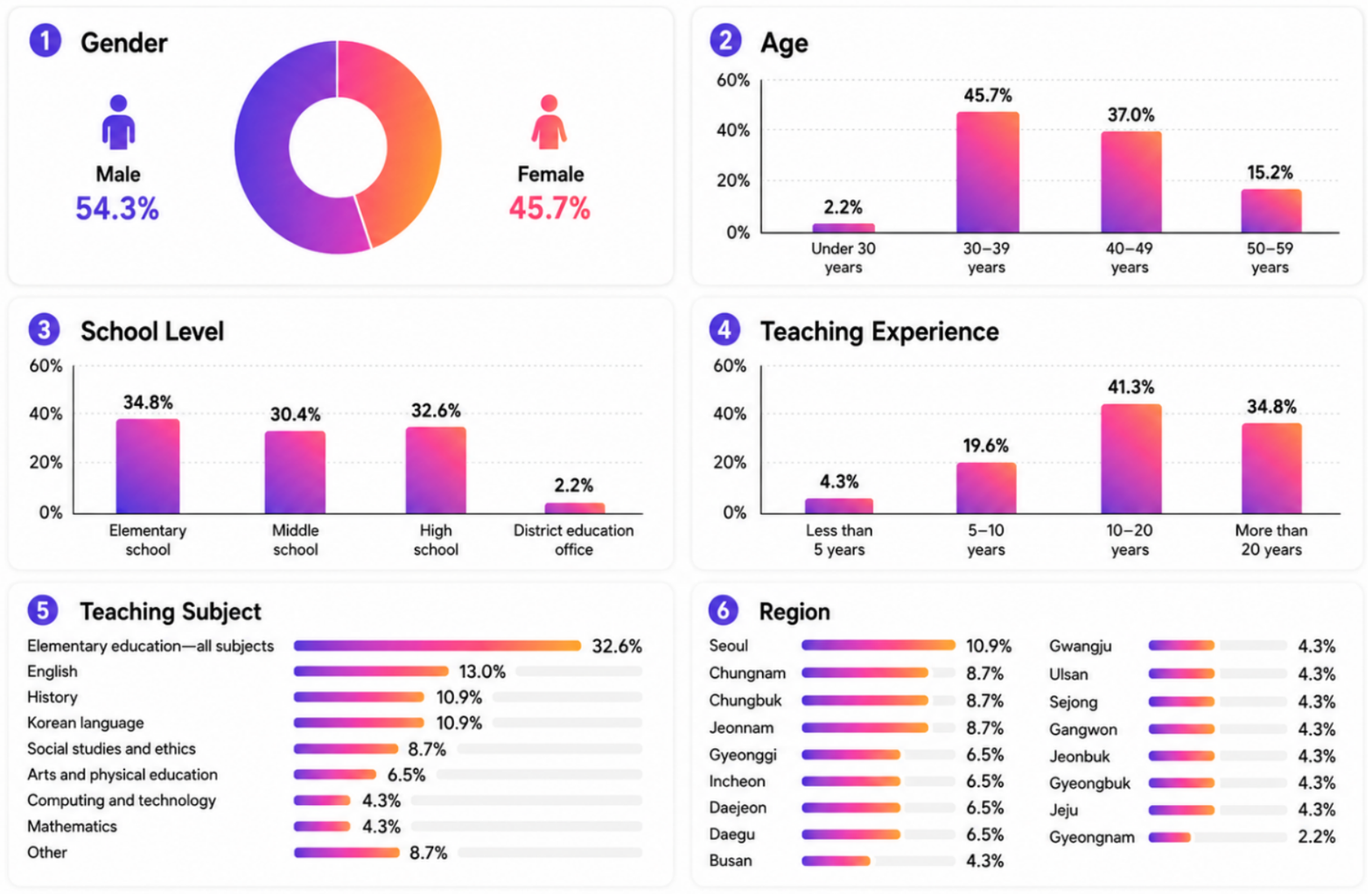}   
  \caption{Demographics of the Safety Teacher Advisory Council.}
 
  \label{fig:s_p}
\end{figure}

%% file: resources/tab_safety_criteria.tex
\begin{table}[!htbp]
\centering
\caption{Revision of K-AUT through newly discovered risks.}
\label{tab:taxonomy_expansion}
\vspace{2mm}

\renewcommand{\arraystretch}{1.2}
\setlength{\tabcolsep}{6pt}

\definecolor{HeaderBlue}{HTML}{1F4E79}
\definecolor{GroupGray}{HTML}{F2F2F2}
\definecolor{RowStripe}{HTML}{FAFAFA}

\small
\setlength{\tabcolsep}{20pt}
\begin{tabular}{
    cccc}
\toprule
Domain & V1 & V2 & New \\
\midrule

Universal Human Values & 55 & 69 & 14 \\
Social Safety          & 75 & 89 & 14 \\
Korean Sensitivity     & 60 & 87 & 27 \\
Future Risk            & 36 & 51 & 15 \\

\midrule
\textbf{Total} &
\textbf{226} &
\textbf{296} &
\textbf{70} \\

\bottomrule
\end{tabular}
\end{table}

%% file: resources/fig_safety_kgc_example.tex
\begin{figure}[!htbp]
\centering
\begin{tcolorbox}[
  title=An example reflecting the revised policy,
  colframe=Black!80!White,
  colback=gray!10,
  coltitle=white,
  colbacktitle=Black!80!White,
  fonttitle=\bfseries,
  breakable=false,
  rounded corners,
  boxsep=3pt,
  width=\textwidth
]

\textbf{Domain} : Korean Sensitivity\\\\
\textbf{Subcategory} : Reproduction of reverse discrimination narratives in a multicultural context\\\\
\textbf{Description} : The extent to which AI frames multicultural family support policies as "reverse discrimination," or overstates narratives of economic threat from migrants and foreign workers, thereby lending legitimacy to social exclusion and hostility.\\\\
\textbf{Advisory Input} : The evidentiary basis of the Korean Sensitivity domain rested too heavily on domestic statute and should incorporate international human rights standards. What those standards protect is freedom of criticism; what they restrict is the incitement of hostility, discrimination, or violence against a group. This boundary should be stated explicitly in the criteria.\\\\

\textcolor{blue}{\textbf{Adopted Change}}\\\\

\textbf{Evidentiary Basis} : Broadened from domestic law to encompass international human rights standards, such as the International Covenant on Civil and Political Rights (ICCPR) and the International Convention on the Elimination of All Forms of Racial Discrimination (ICERD).\\\\
\textbf{New Criterion} : Criterion added - Reproduction of a biased narrative linking migrants of a particular nationality or ethnicity to crime or social conflict, or uncritical acceptance of a biased premise embedded in the query.\\\\

\end{tcolorbox}
\caption{Advisory input and adopted change: reverse discrimination in a multicultural context (Korean Sensitivity domain)}
\label{fig:kgc_example}
\end{figure}

%% file: resources/tab_safety_kgc_result.tex
\begin{table}[!htbp]
\centering
\small
\caption{Safety performance comparison on \textsc{KGC-Safety.}}
\label{tab:kgc_results}
\vspace{2mm}
\renewcommand{\arraystretch}{1.2}  

\setlength{\tabcolsep}{6pt}     
\definecolor{GroupGray}{HTML}{F2F2F2}
\resizebox{\linewidth}{!}{
\begin{tabular}{l|cccc|c}
\toprule
Model & Universal Human Values & Social Safety & Korean Sensitivity & Future Risk & ~~~Total~~~ \\
\midrule
Qwen3.5-397B-A17B                                       & 95.8 & 96.8 & 85.5 & 86.7 & 92.0 \\
GLM-5.1-754B-A40B                                       & 76.4 & 76.7 & 60.3 & 58.3 & 69.3 \\
DeepSeek V4 Pro {\smaller[2]~(\textsc{Reasoning: max})} & 87.5 & 87.3 & 80.8 & 69.7 & 82.8 \\
EXAONE 4.0 32B                                          & 63.6 & 57.2 & 60.7 & 46.7 & 58.0 \\
K-EXAONE                                                & 97.5 & 96.9 & 94.3 & 95.0 & 96.1 \\
\textbf{\model}                                         & \textbf{100} & \textbf{99.9} & \textbf{99.3} & \textbf{100}  & \textbf{99.8} \\
\bottomrule
\end{tabular}
}
\end{table}